%% file: Main.tex
\documentclass[11pt, a4paper]{article}

\usepackage{fullpage}
\usepackage{graphicx} 
\usepackage{amsmath, amsthm}
\usepackage{amssymb}
\usepackage{bbm}
\usepackage{parskip} 
\usepackage[ruled,noend]{algorithm2e}
\usepackage{booktabs}
\usepackage{multirow}
\usepackage{xcolor}
\usepackage{natbib}
\usepackage{enumitem}
\usepackage{url}
\usepackage{array}
\usepackage{comment}
\usepackage{hyperref}
\usepackage{todonotes}
\usepackage[capitalize]{cleveref}

\renewcommand{\algorithmcfname}{ALGORITHM}
\SetAlFnt{\small}
\SetAlCapFnt{\small}
\SetAlCapNameFnt{\small}
\SetAlCapHSkip{0pt}
\IncMargin{-\parindent}

\theoremstyle{plain}
\newtheorem{theorem}{Theorem}[section]
\newtheorem{proposition}[theorem]{Proposition}
\newtheorem{lemma}[theorem]{Lemma}

\theoremstyle{definition}

\theoremstyle{remark}

\newcommand{\red}[1]{{\color{red} #1}}
\newcommand{\cA}{\mathcal{A}}
\newcommand{\cI}{\mathcal{I}}

\newcommand{\OPT}{\text{OPT}}

\newtheorem{claim}[theorem]{Claim}

\newcommand{\mcsdf}{\texttt{MC-SF}}
\newcommand{\mcb}{\texttt{MC-Benchmark}}

\newcommand{\vol}{\textrm{vol}}
\newcommand{\alg}{\mcsdf}
\newcommand{\po}{\tilde{o}}
\newcommand{\xhdr}[1]{\vspace{1mm} \noindent{\bf #1}}
\newcommand{\peak}{\mathrm{peak}}

\title{Online Scheduling for LLM Inference with KV Cache Constraints}

\author{
    Patrick Jaillet\thanks{Department of Electrical Engineering and Computer Science, Massachusetts Institute of Technology. Email: \texttt{jaillet@mit.edu}.}
    \and Jiashuo Jiang\thanks{HKUST. Email: \texttt{jsjiang@ust.hk}.} 
    \and Konstantina Mellou\thanks{Microsoft Research. Email: \texttt{kmellou@microsoft.com}.} 
    \and Marco Molinaro \thanks{Microsoft Research. Email: \texttt{mmolinaro@microsoft.com}.} 
    \and Chara Podimata\thanks{Sloan School of Management, Massachusetts Institute of Technology. Email: \texttt{podimata@mit.edu}.} 
    \and Zijie Zhou\thanks{Operations Research Center, Massachusetts Institute of Technology. Email: \texttt{zhou98@mit.edu}. Part of this work was done during a summer internship at Microsoft Research -- Redmond.}
}

\date{\today}

\begin{document}

\maketitle

\begin{abstract}
Large Language Model (LLM) inference, where a trained model generates text one word at a time in response to user prompts, is a computationally intensive process requiring efficient scheduling to optimize latency and resource utilization. A key challenge in LLM inference is the management of the Key-Value (KV) cache, which reduces redundant computations but introduces memory constraints. In this work, we model LLM inference with KV cache constraints theoretically and propose a novel batching and scheduling algorithm that minimizes inference latency while effectively managing the KV cache's memory. 

More specifically, we make the following contributions. First, to evaluate the performance of online algorithms for scheduling in LLM inference, we introduce a hindsight optimal benchmark, formulated as an integer program that computes the minimum total inference latency under full future information. Second, we prove that no deterministic online algorithm can achieve a constant competitive ratio when the arrival process is arbitrary. Third, motivated by the computational intractability of solving the integer program at scale, we propose a polynomial-time online scheduling algorithm and show that under certain conditions it can achieve a \emph{constant} competitive ratio. We also demonstrate our algorithm's strong empirical performance by comparing it to the hindsight optimal in a synthetic dataset. Finally, we conduct empirical evaluations on a real-world public LLM inference dataset, simulating the Llama2-70B model on A100 GPUs, and show that our algorithm significantly outperforms the benchmark algorithms. Overall, our results offer a path toward more sustainable and cost-effective LLM deployment.
\end{abstract}

\textbf{Keywords:} online optimization, LLM inference, scheduling, competitive ratio

\input{intro}
\input{model}
\input{HO}

\input{Stochastic2}

\input{Numerical}

\input{conclusion}

\bibliographystyle{plainnat}
\bibliography{reference}

\appendix
\input{Preliminary}

\input{Append1}

\input{robustUBApp}

\input{Append4}

\end{document}

%% file: intro.tex
\section{Introduction} \label{sec:intro}

Large Language Models (LLMs) \citep{brown2020language,chowdhery2023palm,openai2023gpt,kaplan2020scaling,wei2022emergent} 
represent a significant advancement in AI, enabling machines to generate human-like text across various languages and contexts. Trained on vast datasets, these models are becoming critical for applications such as chatbots \citep{anthropic2023claude,characterai2023,chatgpt2023,openai2023gpt}, search engines \citep{bing,googlebard,komo,perplexity,you}, code assistants \citep{codewhisperer2023,githubcopilot2023,replitghostwriter2023}, and healthcare services \citep{cascella2023evaluating,peng2023study,sallam2023utility}. 

\noindent {\bf LLM Inference and the KV Cache.} LLMs pose substantial computational challenges, particularly during the inference process where inputs are processed to generate responses. In LLM inference, a \emph{``prompt''} is the input text provided to initiate a model's response generation. These prompts are broken down into smaller units called \emph{``tokens''}, which may consist of words, sub-words, or punctuation marks based on the model's vocabulary.
For instance,
the prompt “What color is the sky?” can be tokenized into six units: “What,” “color,” “is,” “the,” “sky,” and “?”. Similarly, a response like “The color is blue.” would be divided into five tokens: “The,” “color,” “is,” “blue,” and “.”. When a prompt request is processed, typically it is not answered all-at-once; instead, it requires multiples rounds of processing to generate the tokens of the answer sequentially; in the previous example, the output token ``blue'' can only be generated after the preceding one ``is'' is produced by the model.   

Each token is associated with two vectors: the \textbf{Key (K)}, which represents the token's significance to other tokens based on its relevance, and the \textbf{Value (V)}, which stores information that is used in the output if the token is deemed relevant. These KV pairs are computed based \emph{only} on the token content and its absolute position, and once computed, they remain fixed for the entire process.

During inference, the Transformer attention mechanism \citep{vaswani2017attention} uses the stored keys and values to determine how tokens relate to one another. Without optimization, generating each new token would require recalculating the attention scores over all previously seen tokens, leading to a quadratic increase in computation as the sequence grows. To avoid this, modern LLMs use a \emph{KV cache}, which stores all previously computed keys and values. This allows the model to reuse past computations efficiently, reducing the cost of generating each token to linear in the sequence length. However, the KV cache causes memory usage to grow linearly with the number of tokens, and unmanaged growth can result in the GPU running out of memory \citep{hooper2024kvquant,kang2024gear}.

Put together, there are \emph{three} unique challenges posed by LLM inference: {\bf (i)} each token can only be generated after its predecessor; {\bf (ii)} the memory usage grows linearly during inference; and {\bf (iii)} given the scale of real-world LLM inference, decisions must be made within milliseconds, thus making linear or mixed-integer programming solutions unusable.

\begin{figure}[!tb]
\center
\includegraphics[width=0.9\textwidth]{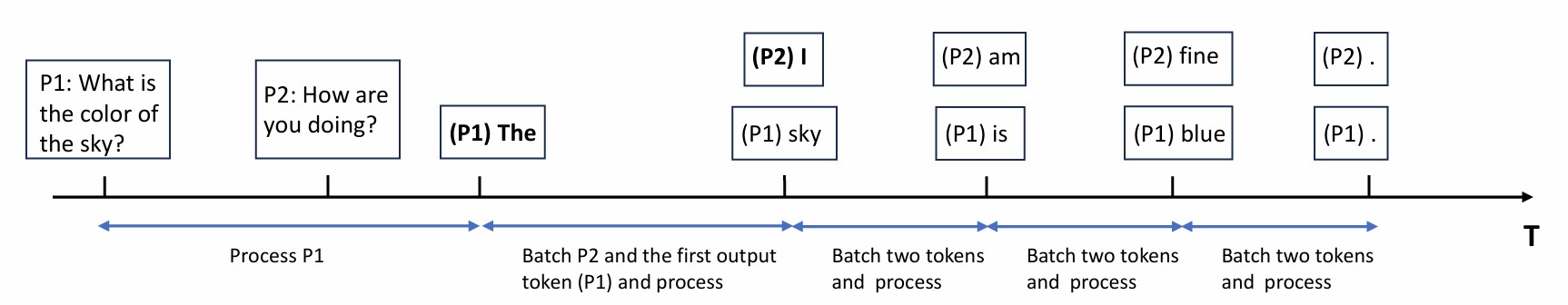}
\caption{Example of online batching and scheduling.} 
\label{fig:example}
\end{figure}

\xhdr{Batching and Scheduling.} When multiple prompt requests are in the queue, batching requests together (rather than handling them one-by-one) and scheduling their processing improves GPU efficiency. For example, Figure \ref{fig:example} illustrates the online batching and scheduling process for two distinct prompts, P1 (``What color is the sky?'') and P2 (``How are you doing?''), during LLM inference {on a single GPU}. Initially, P1 is processed within its own batch. When P2 arrives, it must wait, as simultaneous processing of prompts is limited by worker availability. Once P1 is processed and generates its first output token, ``The,'' the worker batches this token from P1 and the prompt P2 together and processes them together. After P2 produces its first output token, “I”, and P1 produces the next token “sky”, both tokens, “sky” and “I”, are then batched together for efficient token processing, facilitating the subsequent generation of “is” for P1 and “am” for P2.

\xhdr{Our focus.} Our focus in this work is to provide \emph{scheduling} algorithms for LLM inference. Scheduling for LLM inference differs from classical scheduling problems primarily because of the bottlenecks introduced by the KV cache {as we articulated above} (i.e., the linear memory usage growth, and the KV cache's dynamic behavior). {These challenges were also outlined in the recent survey of \citet{mitzenmacher2025queueing}. We also include further explanations of the challenges of scheduling for LLM inference (compared to standard scheduling problems) in Appendix~\ref{subsec:diff}.}

\xhdr{Importance.} Beyond its theoretical importance, optimizing the scheduling policy in LLM inference is crucial in practice for three reasons. First, it can lead to reduced operational costs; the average daily cost of LLM inference for platforms like ChatGPT is approximately \$700,000 \citep{chatgpt_users_2023,sunyan_economics_llm_2023}. Second, it can enhance user satisfaction by minimizing response times. Last but not least, more efficient scheduling promotes sustainability, as LLM inference currently uses vast amounts of electricity and water daily \citep{gordon_2024}. For example, ChatGPT's daily electricity usage exceeds half a million kilowatt-hours (equivalent to the energy consumption of nearly 180,000 U.S. households) and a single conversation uses approximately fifty centiliters of water, akin to the volume of a standard plastic bottle. By optimizing the scheduling policy, we can reduce the number of GPUs required, conserving resources and contributing to sustainability efforts. {In fact, designing environmentally-friendlier LLM inference is an active area of research in systems and engineering, see e.g.,~\citep{li2025ecoserve}.}

\subsection{Results Roadmap}

In this work, we make the following contributions.

{\bf Mathematical model for online batching and scheduling in LLM inference.} While many studies have focused on improving LLM inference efficiency via engineering~\citep{agrawal2023sarathi, kwon2023efficient, patel2023splitwise, pope2023efficiently, sheng2023flexgen, yu2022orca, zhong2024distserve}, there are very few formal models in this space. 
To address this gap, we propose a model (Section \ref{sec:model}) for optimizing the batching and scheduling policy in LLM inference. \emph{Batching} entails selecting which requests to process concurrently, while \emph{scheduling} determines their timing. 

\xhdr{Hindsight optimal benchmark via Integer Programming (IP).} In Section~\ref{sec:HO}, we introduce a hindsight optimal benchmark that assumes complete knowledge of future request arrivals and output lengths. We formulate this as an IP that computes the globally optimal scheduling policy minimizing total end-to-end latency under GPU memory constraints. Our IP captures operational constraints, including non-preemptive execution, per-token memory growth, and memory limits. This optimization benchmark serves as a gold standard for evaluating the quality of online scheduling algorithms.

\xhdr{Online batching and scheduling algorithm.} In Section~\ref{sec:stochastic}, we propose a practical online batching and scheduling algorithm: Memory Constrained Shortest First (\mcsdf). The algorithm prioritizes partially completed requests to reduce latency, and then selects additional waiting requests to fill each batch by maximizing batch size while respecting KV cache memory constraints throughout token generation. We characterize the feasibility of our algorithm by introducing constraints that anticipate future KV cache memory usage, ensuring that the algorithm only schedules batches that remain within memory limits throughout their execution. While we prove that no deterministic online algorithm can in general have bounded \emph{competitive ratio}, 
we show that under some assumptions in the structure of the arrivals, our proposed algorithm has \emph{constant} competitive ratio, 
providing theoretical underpinning for its practical performance. {For our theoretical analysis, we assume that the algorithm has access to (relatively reliable) predictions of the output length of the prompt response; see e.g.,~\citep{zheng2024response} for practical implementation of obtaining such predictions.}

\xhdr{Synthetic-data experiments.} In Section~\ref{subsec:synthetic}, we conduct numerical experiments on synthetically generated instances to evaluate the performance of our proposed \mcsdf \text{ }relative to the hindsight optimal. We design two arrival models to disentangle the sources of performance loss: one where all requests arrive at time zero (eliminating information asymmetry), and one with online stochastic arrivals (reflecting real-world uncertainty). In the first model, \mcsdf \text{ }has nearly optimal performance, with an average latency ratio of 1.005 and \emph{exact} optimality in 114 out of 200 instances. In the second model, where future arrivals are unknown, the average ratio remains competitive at 1.047. 

\xhdr{Experiments on public conversation dataset.} In Section \ref{subsec:realdata}, we perform numerical simulations using the conversation dataset from \cite{zheng2023lmsys}, collected from over 210,000 unique IP addresses via the Vicuna demo and Chatbot Arena website.
We evaluate our algorithm against benchmark parametrized algorithms with six parameter configurations in both high- and low-demand settings. In both the high- and low- demand settings, our algorithm significantly outperforms the benchmark algorithms. These gains can be translated to reduced energy consumption and lower costs, supporting more sustainable and efficient LLM deployment. {Our experimental results in the synthetic and real-world datasets validate the performance of our algorithm beyond the assumptions that we placed in the theoretical part of the work.}

\subsection{Related  Work}

{Our work is primarily related to two streams of literature; online scheduling and LLM inference. Both streams include plethora of papers on variations of the main problem and it is virtually impossible to survey them all in this paper. We include the most relevant and/or representative papers below. Section~\ref{sec:conclusion} includes some related works pertaining to potential future directions stemming from our model.}

\xhdr{Online Scheduling.} In online scheduling, a decision-maker needs to decide the optimal timing for job processing as jobs arrive sequentially. There is a large literature on this subject, where different objectives and input models have been studied; see the books/surveys~\cite{schedulingSurvey,albers2009online,beyondWorstCase,handbookSched}. A particularly relevant set of studies are those that extend beyond processing individual jobs, where jobs of the same type can be grouped into batches and processed simultaneously~\citep{lucier2013efficient,im2013online,liu2015online,li2020online}. Another relevant line of work considers precedence constraints, either within parts of the requests or between different requests~\cite{precSchedAnupam,schedPrecedence,precSchedSchabanel,precSchedBen}. Yet, the unique demands of LLM inference (in particular the growing memory usage in the KV cache and the combination of both batching and the dynamics of sequential token generation) limit the applicability of the existing algorithms. 

\xhdr{LLM Inference {from an Engineering Perspective}.} LLM inference is a developing field with numerous engineering-oriented studies emerging within systems research. For instance, in scheduling, \cite{patel2023splitwise,zhong2024distserve} proposed to use separate GPUs for processing only the prompt and the token phases of a request. 
Regarding batching, works like \cite{yu2022orca,agrawal2023sarathi,agrawal2024taming} examine methods for statically or dynamically grouping pending requests for batch execution. \cite{liu2024deepseek} boosts LLM inference efficiency by introducing multi-head latent attention, which reduces the KV cache through low-rank key-value joint compression. \cite{zhu2023optimal} proposed approximate caching and dynamic choice of model size to accelerate LLM inference. {Another problem in modern LLM inference is head-of-line blocking, where long-running jobs delay shorter ones, thus increasing the average and tail latency. \citet{wu2023fast} tackle this by assigning priority queues for the prompts based on their input length and (re)assigning prompts to different queues as time goes by. Although all of the aforementioned papers boast significant performance gains practically, they are do not come with formal theory bounds. As such, they may be prone to pathological instances. Our approach in this work has been to center the need for theoretical advancements in scheduling for LLM inference.}

{\xhdr{LLM Inference from a Mathematical Perspective.} Recently, there has been a flurry of works from the operations research and optimization community on the theoretical underpinning of scheduling for LLM inference. Similarly to our work,~\citet{bari2025optimal} model LLM serving as an online batching/scheduling problem that must coordinate prompt inputs and responses all while respecting inference-system constraints. Unlike our KV-centric model, they use a more GPU-execution–driven iteration model. \citet{ao2025optimizing} model optimization for LLM inference as a \emph{multi-stage} optimization problem. \citet{li2025throughput} study a model for scheduling in LLM inference centered on throughput/maximal stability, with service times driven by an empirically motivated (piecewise-linear) function of total tokens per batch rather than KV-memory feasibility being the primary constraint as is the case in our paper. Finally, there has also been some recent work on caching for LLM inference~\citep{zhang2025tail}.}

%% file: model.tex
\section{Model} \label{sec:model}

We study a batching and scheduling problem within a discrete time horizon for a single computational worker (GPU). The worker has a memory constraint \( M > 0 \).\footnote{$M$ depends on the memory of the GPU and the complexity of the large language model in use.} Let \( \mathcal{I} \) denote the instance consisting of unprocessed prompt requests arriving over the discrete time horizon. 
Each request \( i \in \mathcal{I}\) has an associated size \(s_i\), representing the number of tokens in the prompt, and response length \( o_i \), indicating the number of tokens in its response.

\xhdr{Request Processing.} Each request is processed online and undergoes two primary phases:

1. \textit{Prompt Phase:} The prompt is processed in order to generate the initial output token. During this phase, the memory required for request \(i\) is \(s_i\), accounting for the storage of key and value matrix for all tokens in the prompt within the KV cache.

2. \textit{Token Phase:} Subsequent tokens are produced sequentially. In this phase, the memory needed to process the \(j\)th token of request \(i\) (\(j \in \{1,2,\ldots,o_i\}\)) is \(s_i + j\). This increment accounts for each new token's key and value, which adds \(1\) to the existing KV cache memory. Consequently, the peak memory usage for processing the final token of request \(i\) reaches \(s_i + o_i\). After the completion of the last token, the KV cache clears all related memory usage \(s_i + o_i\) for that request.

\xhdr{Batch Processing.} A batch may include any unprocessed prompt or output token of different requests; when a prompt request is processed in a batch, its first output token is generated, and, similarly, processing an output token results in the generation of the subsequent token upon batch completion. We assume each batch's processing time is one unit of time, and only one batch can be processed at a time. Moreover, we assume that this process is non-preemptive, meaning that once a prompt request \( i \) is added to a batch, it must be processed continuously for \( o_i \) periods until its final output token is processed. The memory constraint ensures that for all ongoing requests (those not fully processed or pending final output tokens), the total memory usage at any given moment does not exceed \(M\), i.e., if \(S^{(t)}\) is the set of ongoing requests at time $t$ and \(o_i^{(t)}\) is the index of the output token of such request \(i\), then it holds that $\sum_{i \in S^{(t)}} (s_i + o_i^{(t)}) \leq M.$ The scheduler's task is to decide when (i.e., in which batch) to start processing each incoming request. 

\xhdr{Prompt Arrival Process.} 
In this paper, we consider an online arrival model in which unprocessed prompt requests are assigned to the scheduler sequentially over time. An instance \( \mathcal{I} \) consists of prompt requests that arrive one by one, where each request \( i \) is associated with an arrival time \( a_i \); the arrival time \( a_i \), as well as the request sizes \( (s_i,  o_i) \) are revealed only at the moment when request \( i \) arrives. At any time \( t \), the decision-maker has complete knowledge of \( (a_i, s_i, o_i) \) for all requests \( i \) such that \( a_i \leq t \), and no information about requests with \( a_i > t \), which have not yet arrived. {In fact, our proposed algorithm works with only a \emph{prediction} $\po_i \ge o_i$ of the true output sizes; see Section \ref{sec:stochastic} for more details. We remark that methods for high-accuracy output-size prediction are a very active area of research {(e.g.,~\citep{jin2023s,hu2024inference,cheng2024enabling,qiu2024efficient,qiu2024power,fu2024efficient,shahout2024don})}; see also \citet{mitzenmacher2025queueing} for a survey and open problems from a queuing-theory and optimization perspective. For example, \citet{zheng2024response} present a method with prediction accuracy up to $80\%$.}

\xhdr{Evaluation Metrics.} We evaluate an algorithm $\mathcal{A}$'s performance through its \emph{end-to-end latency}. For each request $i$, its end-to-end latency is computed as $c_i(\mathcal A)-a_i$, where $c_i(\mathcal{A})$ is the time the last output token for request $i$ is processed and $a_i$ is the time that request $i$ arrives. We use $\text{TEL}(\mathcal{I}; \mathcal{A}) := \sum_{i \in [n]} c_i(\mathcal A)-a_i$ to denote the total end-to-end latency of algorithm $\mathcal{A}$ for a request sequence $\mathcal{I}$.

%% file: HO.tex
\section{Hindsight Optimal Benchmark and Integer Programming} \label{sec:HO}

To evaluate the performance of an online scheduling algorithm, we introduce a natural benchmark known as the \textit{hindsight optimal}. This benchmark represents an idealized scheduling policy that has complete foresight, i.e., it knows all future request arrivals $a_i$'s and their corresponding prompt and output lengths $s_i$ and $o_i$ at the start of the time horizon. Although such information is not available to any real-world system, the hindsight optimal serves as a gold standard: it represents the best possible performance that any algorithm could achieve given perfect future knowledge.

Define $\bar T$ as an upper bound on the time when all jobs are completed; for instance, we can take $\bar T = \sum_{i \in [n]}(a_i+o_i)$.  We formulate an Integer Program that computes the minimum possible total end-to-end latency achievable by any scheduling policy:
\begin{align}
\min \,& \sum_{i \in [n]} \left(\sum_{t=\{a_i, \dots, \bar T\}} t \cdot x_{i,t}+o_i-a_i \right)  \label{eq:objective} \\
\text{s.t.}\, 
& \sum_{t=\{a_i, \dots, \bar T\}} x_{i,t} = 1, && \forall i \in [n] \label{eq:start_once} \\
&  \sum_{i=1}^{n} \sum_{k=\max\{a_i,t-o_i\}}^{t-1} (s_i+t-k)x_{i,k} \leq M, && \forall t \in [\bar T] \label{eq:memory_constraint} \\
& x_{i,t} \in \{0,1\}, && \forall i \in [n], \forall t \in [\bar T] \label{eq:integrality}
\end{align}

In this formulation, the only decision variable is \( x_{i,t} \in \{0,1\} \), which indicates whether request \( i \) begins processing at time \( t \). Since the system operates under a non-preemptive scheduling regime, each request $i$ must be processed without interruption for \( o_i \) consecutive rounds once it starts. The objective function~\eqref{eq:objective} minimizes the total end-to-end latency over all requests. For each request \( i \), the latency is defined as the time of its last token completion minus its arrival time, i.e., \( (t + o_i) - a_i \), if the request starts processing at time \( t \). This is equivalent to minimizing \( \sum_{t=a_i}^{\bar T} t \cdot x_{i,t} + o_i - a_i  \) across all \( i \), where \( x_{i,t} \) determines the start time.

Constraint~\eqref{eq:start_once} ensures that each request is scheduled exactly once after its arrival. Constraint~\eqref{eq:memory_constraint} enforces the GPU memory limit \( M \) at each time step \( t \), by summing over the memory usage of all requests that are active at that time. Specifically, a request \( i \), if scheduled to start at time \( k \), will be active at time \( t \) for \( k+1 \leq t \leq k + o_i \), which is equivalent to \( k \in [t - o_i, t-1] \). Moreover, since \( i \) cannot be scheduled before its arrival time \( a_i \),  we have \( k \in [\max\{a_i, t - o_i\}, t-1]. \)
If request \( i \) starts at time \( k \), then it contributes to memory usage at time \( t \in [k, k + o_i] \) with an amount equal to \( s_i + t - k \), reflecting both the prompt memory \( s_i \) and the token-wise KV cache growth over time. Lastly, constraint~\eqref{eq:integrality} enforces the binary nature of the scheduling decisions.

This integer program provides an exact characterization of the optimal non-preemptive scheduling policy under complete future information. It jointly captures the timing, memory, and latency structure of the system while remaining compact and interpretable. As such, it serves as a hindsight benchmark for evaluating the performance of online algorithms under both stochastic and adversarial arrival models.

%% file: Stochastic2.tex
\section{Efficient Batching and Scheduling Algorithm and Theoretical Results} \label{sec:stochastic}

In this section, we consider the online arrival model described in Section~\ref{sec:model}, under an adversarial setting where the number of arrivals, the arrival time and size of each prompt, and the output lengths are all chosen by an adversary. 

We start by establishing a hardness result for this general online problem. The standard metric for evaluating the performance of an online algorithm is the \emph{competitive ratio}: an algorithm $\mathcal{A}$ has \emph{competitive ratio $\alpha$} if for every instance $\mathcal{I}$, the algorithm's latency $\text{TEL}(\mathcal{I}; \mathcal{A})$ is at most $\alpha$ times that of the  hindsight-optimal solution $\text{OPT}(\mathcal{I})$.

We show that unfortunately no deterministic online algorithm can achieve a competitive ratio better than order $\sqrt{n}$, i.e., in the worst-case, the gap between any algorithm and the optimal solution needs to grow with the number of requests. The proof of Theorem \ref{thm:advr} can be found in Appendix \ref{app:sto}.

\begin{theorem}
\label{thm:advr}
    Every deterministic algorithm has a competitive ratio at least $\Omega(\sqrt{n})$. 
\end{theorem}

Despite this impossibility result, we propose the scheduling algorithm $\alg$ that has demonstrable effectiveness across a wide range of arrival instances, via both theoretical analysis (below) and numerical experiments (Section~\ref{subsec:synthetic}).

\xhdr{Algorithm $\alg$.} As discussed in Section~\ref{sec:model}, upon the arrival of request $i$, we observe its input length $s_t$ and also a prediction $\po_i$ that overestimates the true output length $o_i$, i.e., $\po_i \ge o_i$. This ensures the algorithm can (over)estimate memory consumption and create feasible batches. 

At each round \( t \), let \( R^{(t)} \) represent the set of all requests that have not yet been processed, while \( S^{(t)} \) denotes the set of requests that are currently in progress but not yet completed (i.e., some output tokens have been generated, but not all of them). Our algorithm prioritizes processing requests in \( S^{(t)} \) first. After processing all the requests currently in $S^{(t)}$, there may still be unused memory in the KV cache, so our algorithm chooses a subset of requests, \( U^{(t)} \subset R^{(t)} \) to add to the batch in order to maximize memory utilization and minimize total latency. To be more precise, our algorithm tries to process as many requests as possible within each batch; for that, it aims to maximize the number of requests in \( U^{(t)} \), provided that they satisfy memory constraints. 

Specifically, for a subset \( U \subset R^{(t)} \), let $t_{\max}(U):= \max_{i\in U} \{ t + \po_i \}$  
represent the maximum predicted completion time for all requests in \( U \) if they are added to the batch at time \( t \). To ensure \( U \) is feasible, the KV cache memory limit must not be exceeded at any \( t' \in [t+1, t_{\max}(U)] \). This requires that:
\begin{equation}\label{eqn:Constraint}
\sum_{i\in S^{(t)}} (s_i+t'-p_i)\cdot \mathbbm{1}_{\{\po_i\geq t'-p_i\}} + \sum_{i\in U} (s_i+t'-t)\cdot \mathbbm{1}_{\{\po_i\geq t'-t\}} \leq M, \quad \forall t' \in [t+1, t_{\max}(U)]
\end{equation}
where \( p_i \) is the starting time to process request \( i \). The first sum accounts for predicted memory usage from ongoing requests in \( S^{(t)} \), while the second captures new requests in \( U \). As long as this inequality is satisfied for all \( t' \in [t+1, t_{\max}(U)] \), \( U \) is feasible to add to the batch. Thus, our selection rule is the following: 1) We sort the set of waiting requests $R^{(t)}$ in non-decreasing order of predicted output lengths $\po_i$'s, and 2) select the largest prefix that satisfies the allowed memory usage imposed by inequality \eqref{eqn:Constraint}:
%
\begin{equation}\label{eqn:Selection}
U^{(t)} = \text{argmax}_{U \textrm{ prefix of $R^{(t)}$ sorted by $\po_i$'s}} \left\{ |U| : \text{inequality \eqref{eqn:Constraint} is satisfied } 
\forall t'\in[t+1, t_{\max}(U)] \right\}
\end{equation}
We prioritize adding requests with smaller $\po_i$ values, as we predict these requests to complete more quickly. Additionally, the predicted peak memory usage of each request is $s_i + \po_i$. In many situations, prompts processed by a single worker tend to have similar input sizes $s_i$, with relatively low variance, while the output length exhibits greater variability due to the stochastic nature of response generation. As a result, $\po_i$ plays a more critical role in determining memory usage, and selecting requests with smaller $\po_i$ values typically reduces peak memory consumption, enabling more requests to be packed into each batch.

 We continue adding requests in this order, checking the feasibility condition of inequality \eqref{eqn:Constraint}. Importantly, we only need to check this constraint at the predicted \textit{completion times} of ongoing or new requests, specifically \( p_j + \po_j \) for \( j \in S^{(t)} \cup U^{(t)} \). This is because (i) memory usage potentially peaks at these completion times, as a request’s memory demand increases until it finishes, and (ii) since memory usage varies linearly between start and end times, satisfying the constraint at these peak points ensures feasibility throughout the interval. The complete algorithm is detailed in Algorithm~\ref{alg:1}. Since this algorithm adds requests to each batch in a shortest-first manner while smartly checking memory constraints over the near future, we refer to it as Memory-Constrained Shortest-First (\mcsdf). 

\begin{algorithm}[htbp] \renewcommand{\algorithmcfname}{Algorithm}
    \SetAlgoLined
    \caption{Memory Constrained Shortest First (\mcsdf)}
    \label{alg:1}
\DontPrintSemicolon
\SetAlgoLined
\For{each round $t = 1$ to $T$}{
    Let $S^{(t)}$ be the set of requests that have already started processing and $R^{(t)}$ be the remaining (waiting) requests at time $t$. Set $U^{(t)} = \emptyset$ \;


    
    \For{each request $i \in R^{(t)}$ in ascending order of predicted output length $\po_i$}{
        Set a list with the times $t' = p_j + \po_j$ for each $j \in S^{(t)}  \cup U^{(t)} \cup \{i\} $\;
        
        \If{all inequalities in Equation~\eqref{eqn:Constraint} hold for all $t'$ in this list}{
            Add request $i$ to $U^{(t)}$\;
        }
        \Else{
            Break the \textbf{for} loop\;
        }
    }
    
    Process the requests in $S^{(t)}  \cup U^{(t)}$\;
    
}
\end{algorithm}

The following proposition states that the computational complexity of each round of \mcsdf \text{ }is actually independent of the number of requests, and the proof can be found in Appendix \ref{app:sto}. 

\begin{proposition}\label{thm:complexity}
    Given that the memory limit of the KV cache is $M$, \mcsdf \text{ }has a computational complexity of $O(M^2)$ at each round $t \in [T]$. 
\end{proposition}

Next, we formally analyze the algorithm $\alg$ for the special case in which all prompts have identical size $s_i = s$\footnote{{A follow-up paper~\citep{wang2025llm} builds on our model and presents methods for dealing with variable input and output prompt lengths.}} and arrive simultaneously at time $t = 0$. This case zooms in on the performance of the algorithm for the important scenario where prompt size variation is small and demand is high, meaning many requests are typically waiting to be processed at any given time. 
Under this setting, we show that \mcsdf \text{ }achieves a constant competitive ratio, long as the predictions $\po_i$ are also within a constant factor of the true output sizes $o_i$.

\begin{theorem}	\label{thm:main}
    Consider instances where all requests arrive at time 0 and have the same prompt size. {Assume the predicted output lengths satisfy $o_i \le \po_i \le \alpha o_i$ for some constant $\alpha \ge 1$}. Then the algorithm $\alg$ is $O(1)$-competitive for such instances, as long as the memory available $M$ is at least twice the maximum predicted memory occupation of a single request in the instance, i.e., $M \ge 2 \max_i (s_i + \po_i)$. 
\end{theorem}

\begin{proof}[Proof of Theorem \ref{thm:main}]
In order to illustrate the key ideas of the analysis, we present here the proof for the case of exact output length predictions, i.e., when $\po_i = o_i$ for all requests $i$. The proof in the presence of prediction errors is identical and just tracks how these errors percolate to the final bound, and details are presented in Appendix \ref{app:stochastic}. We have also not optimized the constants in the $O(1)$, and opted to prioritize simplicity of exposition.

Fix throughout an instance satisfying the assumption of the theorem. Without chance of confusion, we also use $\mcsdf$ to denote the total latency of this algorithm. Also recall that $\OPT$ is the latency of the optimal solution in hindsight. 

For an output length size $o$, let $n_o$ be the number of requests in the instance with this output length. Also, let $\vol_o := s \cdot o + \frac{o \cdot (o+1)}{2}$ denote the volume of memory that a request of this output length $o$ occupies; the first summand $(s \cdot o)$ is because the input tokens $s$ have to stay in the memory until the end of the processing, and the second term is the memory occupied every time that one more token is process until the total $o$ tokens (i.e., $\sum_{j \in [o]} j$ tokens in the memory). We will bound the total latency of $\alg$ and $\OPT$ based on these quantities.

\paragraph{Upper bound on the total latency of $\alg$.}

\begin{lemma}[UB on $\alg$] \label{lemma:UBAlg}
	The total latency incurred by the algorithm $\alg$ is at most
	\begin{align*}
		\frac{1536}{M} \sum_o n_o \cdot \sum_{o' \le o} n_{o'} \cdot  \vol_{o'} + 24 \sum_o n_o \cdot o
	\end{align*}
\end{lemma}

To prove this lemma, we will group the possible output lengths $1,\ldots, o_{\max}$ in powers of 2. More precisely, let $U_\ell$ for $\ell = 0,\ldots, \lfloor \log o_{\max} \rfloor$ denote the set of requests that have output length in the interval $[2^\ell, 2^{\ell + 1})$. Slightly abusing notation, let $\vol_\ell(I)$  be the total amount of memory that the requests $U_\ell$ occupy in $\alg$'s schedule added up over all times in the interval $I$.

\begin{lemma} \label{lemma:UBo}
	Consider one of the sets of requests $U_\ell$, and let $\underline{o}_\ell := 2^{\ell}$ and $\bar{o}_\ell := 2^{\ell + 1} - 1$ denote the smallest and largest possible output length for requests in this set. Let $\underline{t}$ and $\bar{t}$ be the first and last time the algorithm $\alg$ processes a request in $U_{\ell}$. Then the distance between these times can be upper bounded as 
	$$\bar{t} - \underline{t} \,\le\, \frac{192}{M} \cdot \sum_{o = \underline{o}_\ell}^{\bar{o}_\ell} n_o \cdot \vol_o + 5 \bar{o}_\ell.$$
\end{lemma}

\begin{proof}
(For the remainder of the proof we omit the subscript $\ell$ in $\underline{o}_\ell$ and $\bar{o}_\ell$.) To prove this, let us partition the interval $\{\underline{t},\ldots,\bar{t}\}$ into disjoint subintervals $I_1,I_2,\ldots,I_w$ of length $\bar{o}$ (where $I_w$ is the only exceptional interval that can be smaller than $\bar{o}$). For an interval $I \subseteq [T]$, let $\peak(I)$ be the peak memory use by $\alg$ during this interval. The next lemma essentially says that if the peak memory utilization of an interval $I_{j+1}$ is large, then the total volume of requests in $U_\ell$ scheduled ``around'' that point has to also be large.

\begin{claim}[Peak to volume] \label{claim:vol-around}
	Consider any 3 consecutive intervals $I_j, I_{j+1}, I_{j+2}$ of length $\bar{o}$. Then $\vol_\ell(I_j \cup I_{j+1} \cup I_{j+2})\ge \frac{1}{4}\, \peak(I_{j+1}) \cdot \frac{\vol_{\bar{o}}}{s + \bar{o}}$.
\end{claim}

\begin{proof}[Proof of Claim~\ref{claim:vol-around}]
	Let $\tilde{t} \in I_{j+1}$ be the time when the peak memory occupation $\peak(I_{j+1})$ happens. Since the interval $I_{j+1}$ is strictly between times $\underline{t}$ and $\bar{t}$, by definition of the algorithm it only processes requests of $U_\ell$ (i.e., with output lengths between $2^\ell = \underline{o}$ and $2^{\ell + 1} - 1 = \bar{o}$ in $I_{j+1}$), and so only those contribute to the memory occupation at time $\tilde{t}$. If $k$ of these requests contribute to this peak occupation, then each contributes at most $s + \bar{o}$ to it; thus, we have $k \ge \frac{\peak(I_{j+1})}{s+\bar{o}}$. Each such request is completely contained in the bigger interval $I_j \cup I_{j+1}\cup I_{j+2}$, each such request contributes with at least $\vol_{\underline{o}}$ to the memory volume $\vol_\ell(I_j \cup I_{j+1}\cup I_{j+2})$, which then gives $\vol_\ell(I_j \cup I_{j+1}\cup I_{j+2}) \ge k \cdot \vol_{\underline{o}}$. Finally, since $\underline{o} \ge \frac{1}{2} \bar{o}$, a quick calculation shows that $\vol_{\underline{o}} \ge \frac{1}{4} \vol_{\bar{o}}$. Combining these threes inequalities gives the claim. 
\end{proof}

	We now conclude the proof of Lemma \ref{lemma:UBo}.	Suppose for contradiction that $\bar{t} - \underline{t} > \frac{192}{M} \sum_{o = \underline{o}}^{\bar{o}} n_o \cdot \vol_o + 5\bar{o}$. First, since $\bar{o} \le 2 \underline{o}$, a quick calculation shows that $\vol_o \ge \frac{1}{4} \vol_{\overline{o}}$ for all $o$ between $\underline{o}$ and $\bar{o}$. Then since $|U_\ell| = \sum_{o = \underline{o}}^{\bar{o}} n_o$, our assumption implies that $\bar{t} - \underline{t} > \frac{48}{M} \cdot |U_\ell| \cdot \vol_{\bar{o}} + 5\bar{o}$. 
	
	This further implies that $w$ (the number of the intervals $I_j$ of length $\bar{o}$ in this period) is at least 
	\begin{align*}
	w \ge \bigg\lfloor \frac{\frac{48}{M} \cdot |U_\ell| \cdot \vol_{\bar{o}} + 5\bar{o}}{\bar{o}}  \bigg\rfloor \ge \frac{48}{M} \cdot |U_\ell| \cdot \frac{\vol_{\bar{o}}}{\bar{o}} + 4.
	\end{align*}
	Every such interval $I_j$ other than $I_w$ has peak memory utilization more than $M - (s + \bar{o})$, otherwise the algorithm would have scheduled one more request $U_\ell$ in it. Then applying the previous claim to the first $\lfloor \frac{w-1}{3} \rfloor$ groups of 3 consecutive intervals $I_j$'s, we obtain that 
	\begin{align}
	\vol_\ell(\{\underline{t},\ldots,\bar{t}\}) &\ge \bigg\lfloor \frac{w-1}{3} \bigg\rfloor \cdot \frac{1}{4} \cdot (M - (s+\bar{o})) \cdot \frac{\vol_{\bar{o}}}{s+\bar{o}} \nonumber \\
	& \ge \frac{4}{M} \cdot |U_\ell| \cdot \frac{\vol_{\bar{o}}}{\bar{o}} \cdot \frac{M}{2} \cdot \frac{\vol_{\bar{o}}}{s+\bar{o}} > |U_\ell| \cdot \vol_{\bar{o}}, \label{eq:contradiction}
	\end{align}
	where the second inequality uses the assumption $M$ is twice as big as the maximum single request occupation, i.e., $s + \bar{o} \le \frac{M}{2}$, and the last inequality uses $\vol_{\bar{o}} = s \cdot \bar{o} + \frac{\bar{o} \cdot (\bar{o}+1)}{2} > \frac{1}{2} \bar{o} \cdot (s + \bar{o})$. 
	
	However, each request $U_\ell$ contributes at most $\vol_{\bar{o}}$ to $\vol_\ell(\{\underline{t},\ldots,\bar{t}\})$, and thus $\vol_\ell(\{\underline{t},\ldots,\bar{t}\}) \le |U_\ell| \cdot \vol_{\bar{o}}$; this contradicts Equation~\eqref{eq:contradiction}, and concludes the proof of the lemma. 	
\end{proof}

\begin{proof}[Proof of Lemma \ref{lemma:UBAlg}]
	Let $t_\ell$ be the first time a request in $U_\ell$ is (starting to be) processed by $\alg$ (let $\ell_{\max} := \lfloor \log o_{\max} \rfloor$ and let $t_{\ell_{\max} + 1}$ be the last time the algorithm is processing something). The latency for each request in $U_\ell$ is at most {$t_{\ell+1} + \bar{o}_\ell$} (i.e., if the algorithm starts processing requests from the next group then is has already started to process all requests in $U_\ell$, which take at most $+ \bar{o}_\ell$ time to complete); thus, the total latency is at most $\sum_\ell |U_\ell| \cdot ({t_{\ell+1} + \bar{o}_\ell})$. However, from the Lemma~\ref{lemma:UBo} we know that $$t_{\ell+1} - t_{\ell} \,\le\, \frac{192}{M} \cdot |U_\ell| \cdot \vol_{\bar{o}_\ell} + 5\bar{o}_\ell + 1 \,\le\, \frac{192}{M} \cdot |U_\ell| \cdot \vol_{\bar{o}_\ell} + 6\bar{o}_\ell,$$ and so $t_{\ell + 1} \le \frac{192}{M} \sum_{\ell' \le \ell} |U_{\ell'}| \cdot \vol_{\bar{o}_{\ell'}} + 6 \sum_{\ell' \le \ell} \bar{o}_{\ell'}$.
    
    This gives that the total latency of $\alg$ can be upper bounded as 
	\begin{align}
		\alg \,\le\, \frac{192}{M} \underbrace{\sum_\ell |U_\ell| \sum_{\ell' \le \ell} |U_{\ell'}| \cdot \vol_{\bar{o}_{\ell'}}}_{A} + 6 \underbrace{\sum_\ell |U_\ell| \sum_{\ell' \le \ell} \bar{o}_{\ell'}}_{B}.  \label{eq:UBAlg}
	\end{align}
	Concluding the proof of the lemma requires just a bit of algebra to clean up the bound. 
	
	To upper bound the term $A$, let $O_\ell := \{\underline{o}_\ell, \ldots, \bar{o}_\ell\}$ be the possible output lengths of the requests in $U_\ell$. We first observe that since $\bar{o}_{\ell} \le 2\underline{o}_\ell$, we have $\vol_{\bar{o}} \le 4 \vol_o$ for every $o \in O_\ell$. Moreover, since $|U_\ell| = \sum_{o \in O_\ell} n_o$, we have 
	\begin{gather*}
	|U_\ell|^2 \cdot \vol_{\bar{o}_\ell} \le 2 \bigg(\sum_{o \in O_\ell} n_o \sum_{o' \in O_\ell, o' \le o} n_{o'}\bigg) \cdot  \vol_{\bar{o}_\ell} \le  8 \sum_{o \in O_\ell} n_o \sum_{o' \in O_\ell, o' \le o} n_{o'} \cdot  \vol_{o'}
	\end{gather*}
	and for $\ell' < \ell$ 
	\begin{gather*}
	|U_\ell| \cdot |U_{\ell'}| \cdot \vol_{\bar{o}_{\ell'}} \le \bigg(\sum_{o \in O_\ell} n_o \sum_{o' \in O_{\ell'}} n_{o'}\bigg) \cdot  \vol_{\bar{o}_{\ell'}} \le  4 \sum_{o \in O_\ell} n_o \sum_{o' \in O_{\ell'}} n_{o'} \cdot  \vol_{o'},
	\end{gather*}	
	which combined give
	\begin{gather*}
	|U_\ell| \sum_{\ell' \le \ell} |U_{\ell'}| \cdot \vol_{\bar{o}_{\ell'}} \,\le\, 8  \sum_{o \in O_\ell} n_o \sum_{o' \le o} n_{o'} \cdot  \vol_{o'},
	\end{gather*}	
	and so adding up over all $\ell$ gives the upper bound $A \le 8 \sum_o n_o \sum_{o' \le o} n_{o'} \cdot  \vol_{o'}.$
	
	To upper bound the term $B$ in \eqref{eq:UBAlg}, we observe that since the $\bar{o}_\ell$'s grow exponentially, $\sum_{\ell' \le \ell} \bar{o}_{\ell'} \le 2 \bar{o}_{\ell}.$ Then since $\bar{o}_{\ell} \le 2 o$ for every $o \in O_{\ell}$, we have
	\begin{align*}
		B \le 2 \sum_\ell |U_\ell| \cdot \bar{o}_{\ell} = 2 \sum_\ell \sum_{o \in O_\ell} n_o \cdot \bar{o}_{\ell} \le  4 \sum_\ell \sum_{o \in O_\ell} n_o \cdot o = 4 \sum_o n_o \cdot o.
	\end{align*}
	Plugging these upper bounds on $A$ and $B$ on \eqref{eq:UBAlg}, we get
	\begin{align*}
		\alg \le \frac{1536}{M} \sum_o n_o \sum_{o' \le o} n_{o'} \cdot  \vol_{o'} + 24 \sum_o n_o \cdot o.
	\end{align*}
	This concludes the proof of Lemma \ref{lemma:UBAlg}.
\end{proof}


\paragraph{Lower bound on the total latency of $\OPT$.} 

\begin{lemma} \label{lemma:LBOPT}
	We have the following lower bound on the total optimal latency $\OPT$:
	\begin{align*}
		\OPT \ge \frac{1}{6M} \sum_o n_o \cdot \sum_{o' \le o} n_{o'} \cdot \vol_{o'} + \frac{1}{6} \sum_o n_o \cdot o.
	\end{align*}
\end{lemma}

To lower bound the total latency of $\OPT$ we consider the following LP relaxation. Let $U_o$ denote the set of requests with output length $o$. Let $\bar{a}^t_o$ be the number of requests in $U_o$ that finish at time $t$ in the optimal solution. The memory volume of all requests that finish up to time $t$ need to fit in the total memory $t \cdot M$ available up to that time, and hence $\sum_{t' \le t} \sum_o \bar{a}^t_o \cdot \vol_o \le t \cdot M$. Moreover, $\sum_t \bar{a}^t_o = n_o$ (all requests in $U_o$ finish at some time). Finally, the $\sum_o \bar{a}^t_o$ requests that finish at time $t$ have latency (recall all requests are released at time 0) equal to $t$, and the optimal latency $\OPT$ is given by $\sum_t t \cdot \sum_o \bar{a}^t_o$. Together these observations show that $\OPT$ can be lower bounded by the following Linear Program with variables $a^t_o$, where in particular we relax the requirement that $\bar{a}^t_o$'s are integers:
\begin{align}
	\OPT_{LP} := \min &\sum_t t \cdot \sum_o a_o^t \notag\\
	\textrm{s.t.}\,& \sum_{t' \le t} \sum_o a^t_o \cdot \vol_o \le t \cdot M, \qquad \forall t \label{eq:LPmem}\\
	&\sum_t a^t_o = n_o, \qquad \forall o \notag\\
	&a^t_o \ge 0, \qquad \forall t,o. \notag
\end{align}

We then lower bound $\OPT_{LP}$. Consider the optimal solution $\{a^{* t}_o\}_{t,o}$ for this LP. For a given output size $o$, let $t^*_o$ be the first time $t$ where $a^{* t}_o > 0$, i.e., where a request $U_o$ is assigned to time $t$. 

Indeed we can see the above LP as the problem of, for each $o$, (fractionally) assigning $n_o$ units over the timesteps $0,1,\ldots$ subject to the constraint that the timesteps have a limited receiving capacity, and assigning a unit to time $t$ incurs a cost of $t$. Since $\vol_o$ is increasing over $o$, we observe that the optimal solution to this LP is to first try to assign all requests with the smallest $o$ (call it $o_{\min}$) to time 1; if the constraint $\sum_o a^1_o \cdot \vol_o \le M$ does not allow all $n_{o_{\min}}$ requests to be assigned to time 1, the remaining ones are assigned to time 2, and so on; otherwise we move to the 2$nd$ smallest $o$ and assign the requests $U_o$ to time 1. This optimizes the above LP because it maximizes the number of items assigned to time 1, which has the smallest cost (the analogous argument holds for the other times). In summary, we have that the ``first time'' values $t^*_o$ are non-decreasing, namely $t^*_o \le t^*_{o'}$ when $o < o'$. 

Let $t^*_{o_{\max} + 1}$ be the last time such that $\sum_o a^{* t}_o > 0$. Using this observation we will prove the following lower bound on the ``first times'' $t^*_o$.

\begin{claim}	\label{claim:tStar}
	For all $o$ we have $t^*_{o + 1} \ge \frac{1}{M} \sum_{o' \le o} n_{o'} \cdot \vol_{o'}$. 
\end{claim}

\begin{proof}
	By the observation above, in the optimal solution $\{a^{*o'}_t\}_{o',t}$ all items with output length at most $o$ are assigned to times $\le t^*_{o + 1}$, i.e., no later than when the next output length is assigned; thus, $\sum_{t' \le t^*_{o + 1}} a^{*t'}_{o'} = n_{o'}$ for all $o' \le o$. Then considering \eqref{eq:LPmem} to time $t^*_{o + 1}$ we get
	\begin{align*}
		\sum_{o' \le o} n_{o'} \cdot \vol_{o'} = \sum_{t' \le t^*_{o + 1}} \sum_{o' \le o} a^{*t'}_{o'} \cdot \vol_{o'} \le t^*_{o + 1} \cdot M,
	\end{align*}
	and rearranging we get the claim. 
\end{proof}

We are now able to prove the lower bound on $\OPT$ from Lemma \ref{lemma:LBOPT}.

\begin{proof}[Proof of Lemma \ref{lemma:LBOPT}]
	By definition of $t^*_o$, we know that $a^{*t}_o = 0$ for all $t < t^*_o$, and so $\sum_t t \cdot a^{*t}_o \ge t^*_o \cdot \sum_{t \ge t^*_o} a^{*t}_o = t^*_o \cdot n_o$. Plugging the bound on $t^*_o$ from the previous claim and adding over all $o$ we can lower bound $\OPT_{LP}$ (and thus $\OPT$) as
	\begin{align}
	\OPT \ge \OPT_{LP} &= \sum_t t \cdot \sum_o a^{*t}_o \notag\\
	&\ge \sum_o n_o \cdot t^*_o \notag\\
	&\ge \sum_o n_o \cdot \bigg( \frac{1}{M} \sum_{o' < o} n_{o'} \cdot \vol_{o'} \bigg) \notag\\
	&=  \frac{1}{M} \sum_o n_o \sum_{o' \le o} n_{o'} \cdot \vol_{o'} - \frac{1}{M} \sum_o n_o^2 \cdot \vol_o.\label{eq:LBOPT1}
	\end{align}
	To remove the negative term in the right-hand side (and add a new one, to match that of Lemma \ref{lemma:UBAlg}), we provide two other lower bounds on the original $\OPT$ (not the LP). 
	
	The first is that for any output length $o$, due to memory constraints, at most $\frac{n_o}{2}$ of them can be finished in the optimal schedule by time $\frac{n_o}{2} \frac{\vol_o}{M}$, since the total memory available up to this time is $\frac{n_o}{2} \cdot \vol_o$ and each such request consumes $\vol_o$ memory; thus, the total latency on the optimal solution for the request of output length $o$ is at least $\frac{n_o}{2} \cdot \frac{n_o}{2} \frac{\vol_o}{M}$. Adding this over all $o$, the total latency $\OPT$ has the lower bound 
	\begin{align}
	\OPT \ge \frac{1}{4M} \sum_o n^2_o \cdot \vol_o.  \label{eq:LBOPT2}
	\end{align}	
	Finally, since each request of output length o takes $(o+1)$ units of processing/time to finish (and thus has latency at least $o$), we also have $\OPT \ge \sum_o n_o \cdot o$ (which is at least $\sum_o n_o$).
	
	Adding this bound plus 4 times \eqref{eq:LBOPT2} plus \eqref{eq:LBOPT1} we get 
	\begin{align}
	6\, \OPT \ge  \frac{1}{M} \sum_o n_o \sum_{o' \le o} n_{o'} \cdot \vol_{o'} + \sum_o n_o \cdot o.
	\end{align}	 
	This concludes the proof of Lemma \ref{lemma:LBOPT}.
\end{proof}


Combining the bound on $\alg$ from Lemma \ref{lemma:UBAlg} and the lower bound on $\OPT$ from Lemma \ref{lemma:LBOPT}, we obtain $\alg \le O(1) \cdot \OPT$, thus proving Theorem \ref{thm:main}.
\end{proof}

%% file: Numerical.tex
\section{Numerical Simulation} \label{sec:num}

In this section, we conduct numerical simulations using both synthetic data and real-world traces to evaluate the performance of our proposed scheduling algorithm \mcsdf. 

In the first of these experiments (Section~\ref{subsec:synthetic}), the primary goal of  is to assess the performance of \mcsdf \text{ }within the framework of the mathematical model introduced in Section~\ref{sec:model}, comparing it against the hindsight-optimal policy from Section~\ref{sec:HO}.
The second set of experiments (Section~\ref{subsec:realdata}) 
evaluates the practical effectiveness of \mcsdf \text{ }using a large-scale, open-source dataset derived from real LLM inference traces. Due to the size and complexity of this dataset, it is computationally infeasible to compute the hindsight optimal-solution, and we instead compare \mcsdf \text{ }against several baseline scheduling policies that are representative of those used in real-world LLM inference systems. 
Additional experimental results are provided in Appendix~\ref{app:num}.

\subsection{Synthetic Data Numerical Simulations} \label{subsec:synthetic}

We first evaluate the performance of \mcsdf \text{ }on synthetic datasets. These experiments aim to quantify the two primary sources of performance gap between the algorithm and the hindsight optimal: (i) the information asymmetry between online and offline algorithms, and (ii) the inherent suboptimality of the algorithm itself. To partially disentangle these effects, we consider two different arrival models{, which we describe below}. To obtain the hindsight-optimal solution, we solve the integer programming model from Section~\ref{sec:HO} using the Gurobi solver. Since solving the Integer Program is computationally expensive, we limit this comparison to small-scale synthetic instances to meet reasonable memory and time limits.

\paragraph{Experimental Setup.} For each experiment, the memory capacity \( M \) is drawn uniformly from the integers between 30 and 50. For each request, the prompt size \( s_i \) is drawn uniformly from the integers between 1 and 5, and the output length \( o_i \) from the integers between 1 and \( M - s_i \).
We run 200 independent trials under each of the following arrival models:

1. \textit{Arrival Model 1 (All-at-once arrivals).} All \( n \) requests arrive at time \( t = 0 \), where \( n \) is a random integer between 40 and 60. This model allows both \mcsdf \text{ }and the hindsight optimal to access the full set of requests at the beginning of execution, eliminating any information gap. Therefore, any observed performance difference is solely due to the algorithm’s structural approximation.
    
2. \textit{Arrival Model 2 (Online stochastic arrivals).} Requests arrive over a discrete time horizon \( [1, T] \), where \( T \) is drawn uniformly from the integers between 40 and 60. The arrival process follows a stationary Poisson distribution with rate \( \lambda \in [0.5, 1.5] \), sampled uniformly. This now introduces \emph{online uncertainty}, which the algorithm must respond to without knowing future arrivals.

\xhdr{Performance Metric.} We evaluate performance using the ratio between the total end-to-end latency incurred by \mcsdf \text{ }and that of the hindsight-optimal policy.
A ratio of 1 indicates that the algorithm achieves the optimal schedule, while larger values reflect a performance gap. 

\begin{figure}[!tb]
\center
\includegraphics[width=0.48\textwidth]{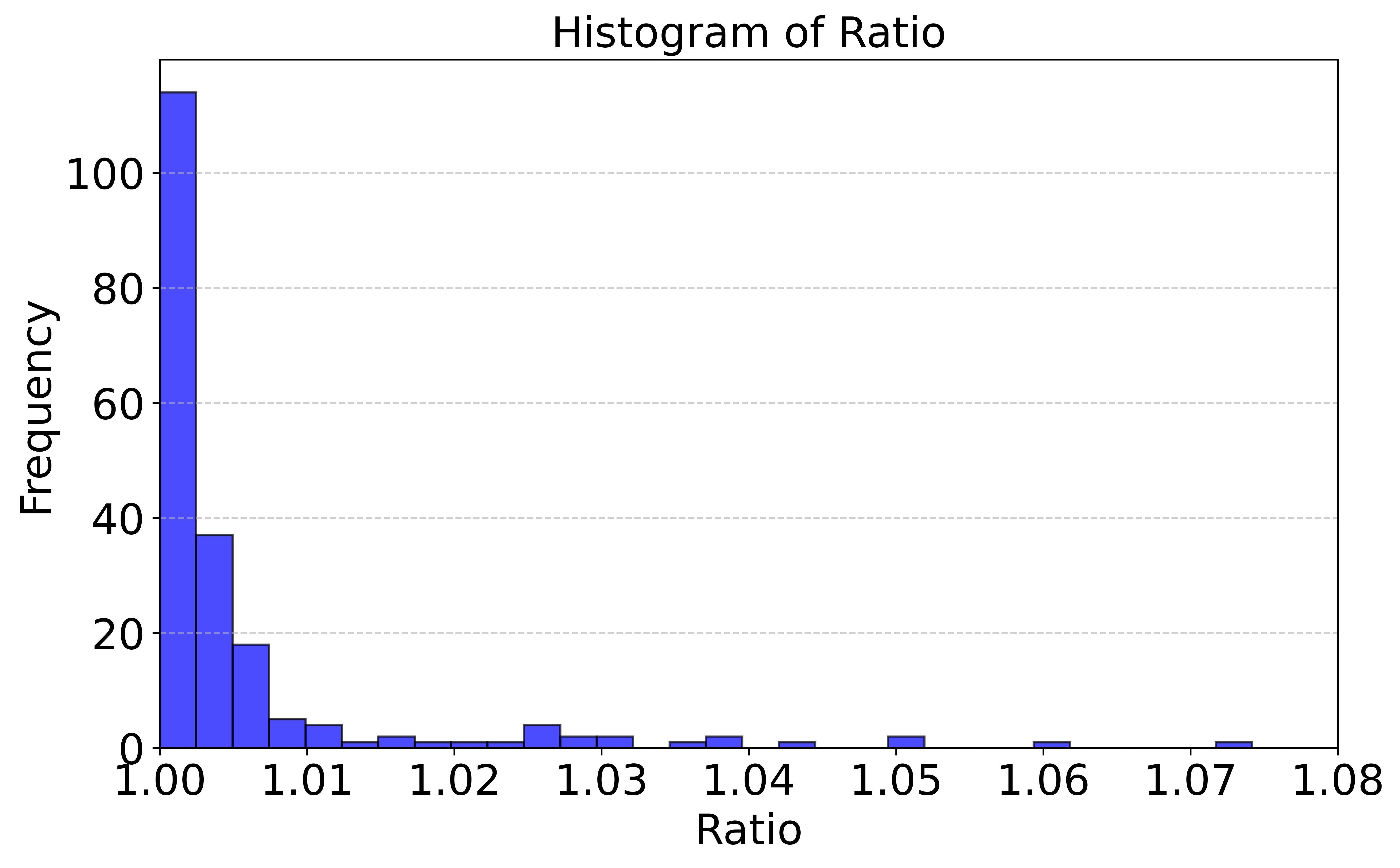}
\includegraphics[width=0.48\textwidth]{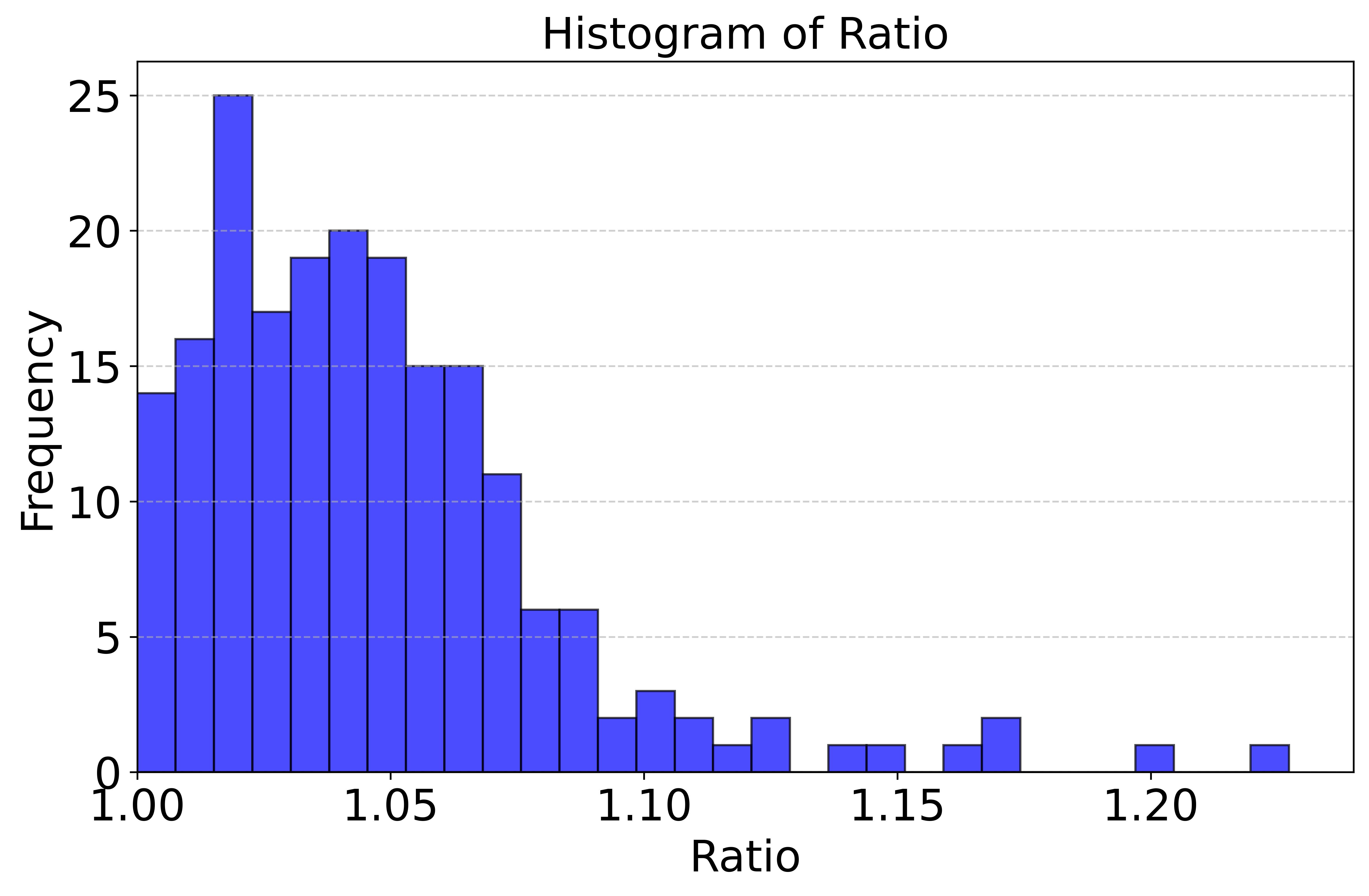}
\caption{Histogram of latency ratio: \mcsdf \text{ }vs Hindsight Optimal. \textbf{Left: }Arrival Model 1. \textbf{Right: }Arrival Model 2.} 
\label{fig:histograms}
\end{figure}

\xhdr{Results for Arrival Model 1.} Across 200 trials, the best-case ratio is 1.000 (meaning the algorithm produces the optimal schedule), the worst-case ratio is 1.074, while the average ratio is 1.005. The left figure of Figure \ref{fig:histograms} displays the distribution of latency ratio of \mcsdf \text{ }vs. Hindsight Optimal under Arrival Model 1. These results show that in this setting, \mcsdf \text{ }performs almost optimally. In 114 of the 200 trials, the algorithm exactly matches the optimal latency, demonstrating that its design closely approximates the structure of the hindsight-optimal policy. Since in this first arrival model there is no information gap between \mcsdf~ and hindsight-optimum, both having full knowledge of the input, these ratios close to 1 indicate that our proposed algorithm has a very small structural suboptimality.

\xhdr{Results for Arrival Model 2.} 
Across 200 trials, here the best-case ratio is 1.000, the worst-case ratio is 1.227, and the average ratio is 1.047. The right figure of Figure \ref{fig:histograms} displays the distribution of latency ratio of \mcsdf \text{ }vs. Hindsight Optimal under Arrival Model 2.  Since 
{the requests} are no {longer} known a priori by the algorithm, the performance ratio captures both the effect of information asymmetry and the algorithm's structural approximation error. The resulting ratios are still quite close to 1, indicating strong performance of \mcsdf \text{ }and establishing it as an effective and practically viable scheduling policy. The difference of the ratios between the two models can be partially attributed to the cost of online decision-making.

\subsection{Real Data Numerical Experiments} \label{subsec:realdata}

\xhdr{Dataset Overview.} We use a conversational dataset\footnote{Publicly available at \url{https://huggingface.co/datasets/lmsys/lmsys-chat-1m}.} by \cite{zheng2023lmsys} from over 210,000 distinct IP addresses via the Vicuna demo and Chatbot Arena platforms. To manage its size, we selected a random subset of 10,000 conversations for analysis. Each conversation includes a user-generated question and a response from an LLM. We define the question as the prompt input and each word in the response as an output token. 
Figure \ref{fig:distribution} in Appendix~\ref{app:num} shows the distribution of the sizes (word count) of prompts (mean: $40.62$ and median: $11$) and output tokens (mean: $85.32$ and median: $45$).

\xhdr{Simulation Setup.} In our simulation, we operate over a continuous time horizon, with 10,000 prompts arriving according to a stationary Poisson process. Let $\lambda$ denote the arrival rate per second, where we consider two cases: \textbf{Case 1: High Demand $\lambda=50$} and \textbf{Case 2: Low Demand $\lambda=10$}. In both scenarios, we simulate the performance of the Llama2-70B model on two linked A100 GPUs (operating collectively as a single worker) and consider the memory limit $M = 16492$; additional details can be found in Appendix~\ref{app:num}.  

At a high level, the simulation works as follows: prompts chosen from the conversational dataset arrive continuously via the chosen Poisson arrival process. To calculate the processing time of a batch created by an algorithm, we use {an} LLM inference simulator from \cite{agrawal2024vidur}.\footnote{Publicly available at \url{https://github.com/microsoft/vidur/}.}
We then report {the} average end-to-end latency,
i.e., the total end-to-end latency divided by the number of requests.

\xhdr{Benchmark Algorithms.} Currently, most LLM inference research is built on vLLM, a widely recognized high-performance inference engine designed to enhance the efficiency of Large Language Model (LLM) inference \citep{kwon2023efficient}. Since LLM inference is a relatively new field, scheduling algorithms in both academia and industry primarily rely on those implemented in vLLM. The scheduling algorithm in vLLM follows a first-come, first-served (FCFS) strategy: when the machine is idle, it prioritizes requests based on their arrival time. However, instead of maximizing the number of requests in a batch, vLLM uses a predefined threshold for memory occupation. Once the KV cache occupancy exceeds this threshold, no additional requests are added to the batch. This motivates us to define the following benchmarks:

\textit{$\alpha$-protection greedy algorithms.} We begin with a class of parameterized algorithms called $\alpha$-protection greedy scheduling algorithms, where $\alpha \in (0,1)$. These algorithms maintain a protection memory threshold of $\alpha M$, serving as a safeguard against memory overflow. When forming each batch, the algorithm gives priority to existing token jobs. For new prompt requests, it checks if adding a new prompt $i$ with initial memory $s_i+1$ would cause the memory usage to exceed $(1-\alpha)M$. If the memory limit is exceeded, no further prompts are added to the batch. 
In the event that the KV cache memory overflows during the processing of the requests, the $\alpha$-protection greedy scheduling algorithms will clear all active requests sending them back to the waiting queue as unprocessed.

\textit{$\alpha$-protection, $\beta$-clearing algorithms.} $\alpha$-protection greedy scheduling algorithms can cause unnecessary evictions; to address this, we define a new class of benchmark algorithms: $\alpha$-protection $\beta$-clearing algorithms. These algorithms follow the same principles as $\alpha$-protection greedy scheduling but, when the KV cache memory limit is exceeded, each active request is cleared and sent back to the scheduler with an independent probability $\beta$.

\textit{Memory Constrained Benchmark (\mcb).} Beyond the benchmarks discussed above, we introduce an additional benchmark algorithm, denoted as \mcb, which is partially inspired by the vLLM scheduling policy and partially by the memory feasibility checks used in \mcsdf. Following the vLLM structure, \mcb~forms batches by processing requests in ascending order of their arrival times. While constructing a batch, it decides whether to include each request based on a prospective memory check used in \mcsdf: it only adds a request if the future memory usage, including KV cache growth, remains within the memory limit \( M \). The detailed pseudocode for \mcb~is provided in Appendix~\ref{app:num} (Algorithm \ref{alg:2}).

\subsubsection{Results: High vs. Low Demand}

We evaluate the performance of \mcsdf{} and baseline algorithms under two demand regimes: high demand ($\lambda = 50$) and low demand ($\lambda = 10$).

First, we examine memory usage over time. Figures~\ref{fig:movertime} and~\ref{fig:Lmovertime} show that \mcsdf{} consistently stays within the memory limit $M$ across both settings. Despite batch processing durations varying in practice, the memory check in Equation~\eqref{eqn:Selection} reliably prevents overflow by ensuring feasibility over future memory usage. Under low demand, memory usage remains close to full utilization, suggesting a stable system state.

We also investigate the performance of the $\alpha$-protection and $\alpha$-protection–$\beta$-clearing heuristics.
We observe that for very small protection levels $\alpha$, the $\alpha$-protection heuristic may lead to repeated evictions and infinite processing loops, therefore we perform a grid search with step size $0.01$ to identify the smallest possible $\alpha$: $\alpha = 0.21$ in the high-demand setting and $\alpha = 0.24$ in the low-demand setting.
We then evaluate latency across six configurations, including $\alpha = 0.3$, $\alpha = 0.25$, and four combinations of $\alpha \in \{0.2, 0.1\}$, $\beta \in \{0.2, 0.1\}$.

\begin{figure}[!tb]
\center
\includegraphics[width=0.48\textwidth]{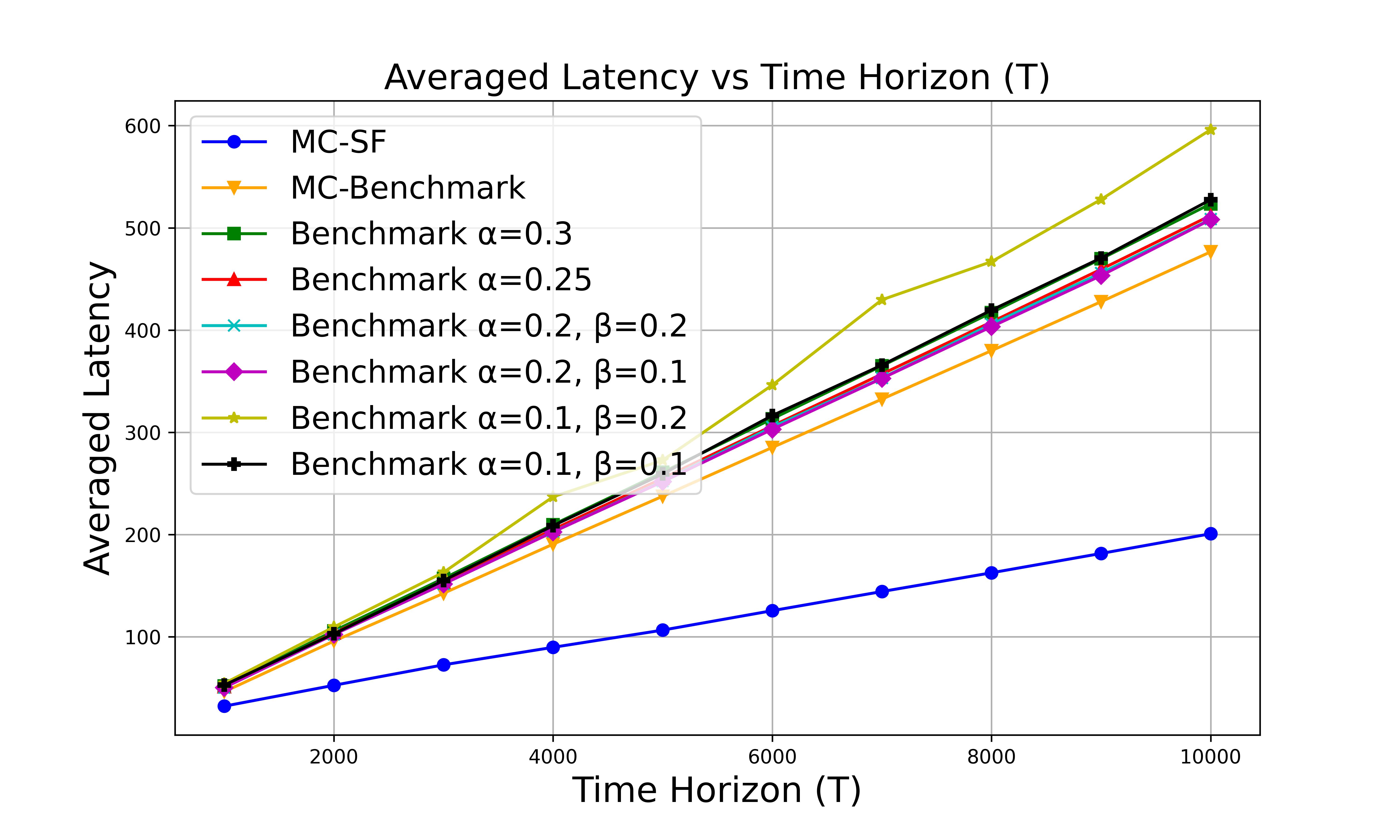}
\includegraphics[width=0.48\textwidth]{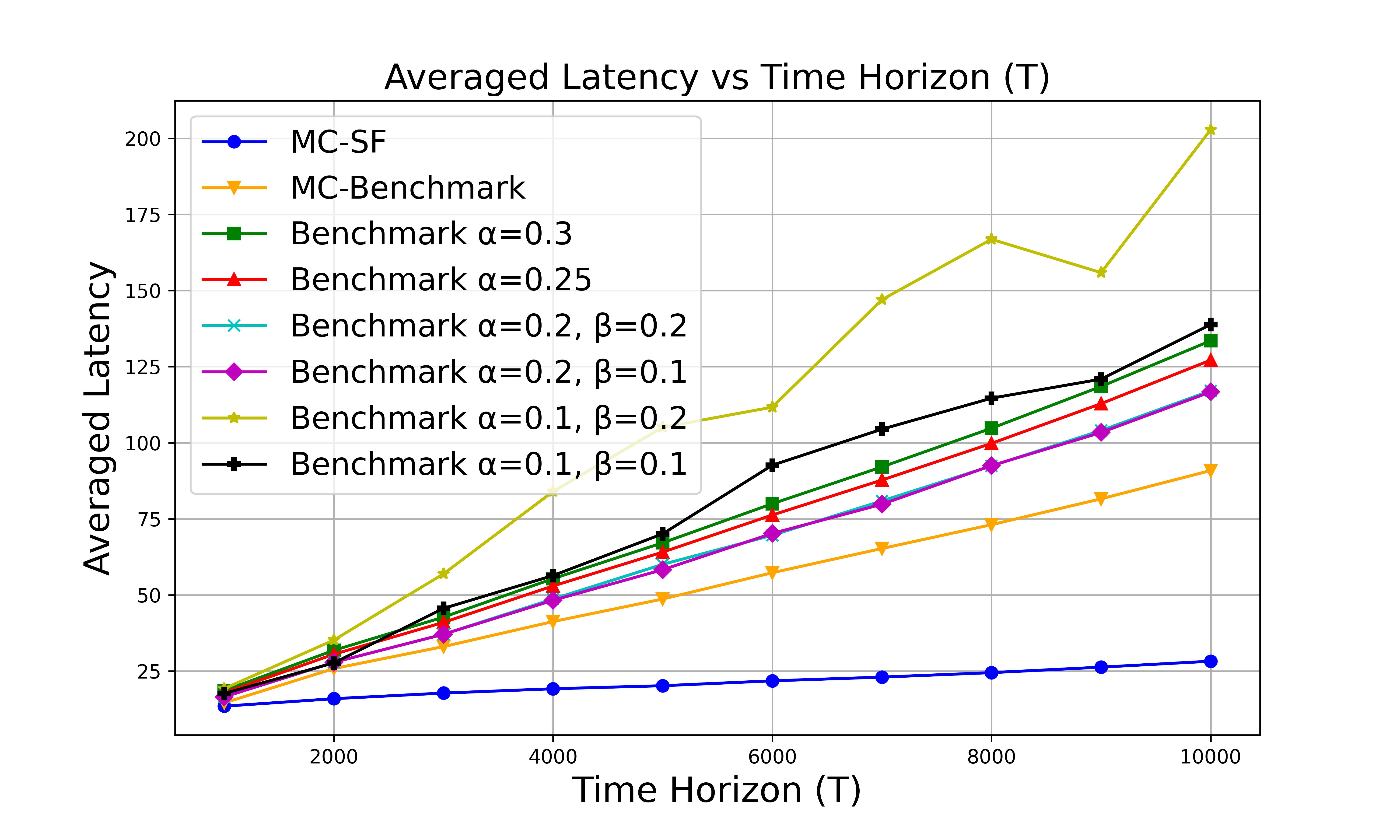}
\caption{Average End-to-End Latency Across Scheduling Algorithms. \textbf{Left: }High Demand. \textbf{Right: }Low Demand.} 
\label{fig:latencyhigh}
\end{figure}

Figure~\ref{fig:latencyhigh} presents the average end-to-end latency across all algorithms for request volumes $\{1000, $ $ 2000, \ldots, 10000\}$. In the high-demand case (left), average latency increases linearly for all algorithms, indicating overload. Notably, \mcsdf{} has a slope of approximately $1/6$, compared to $1/2$ for the best-performing benchmark, highlighting its superior scalability. In the low-demand case (right), \mcsdf{} maintains a much lower latency growth rate—about $1/800$, which is over eight times smaller than the best benchmark slope ($\sim 1/100$).

\begin{figure}[!tb]
\center
\includegraphics[width=0.6\textwidth]{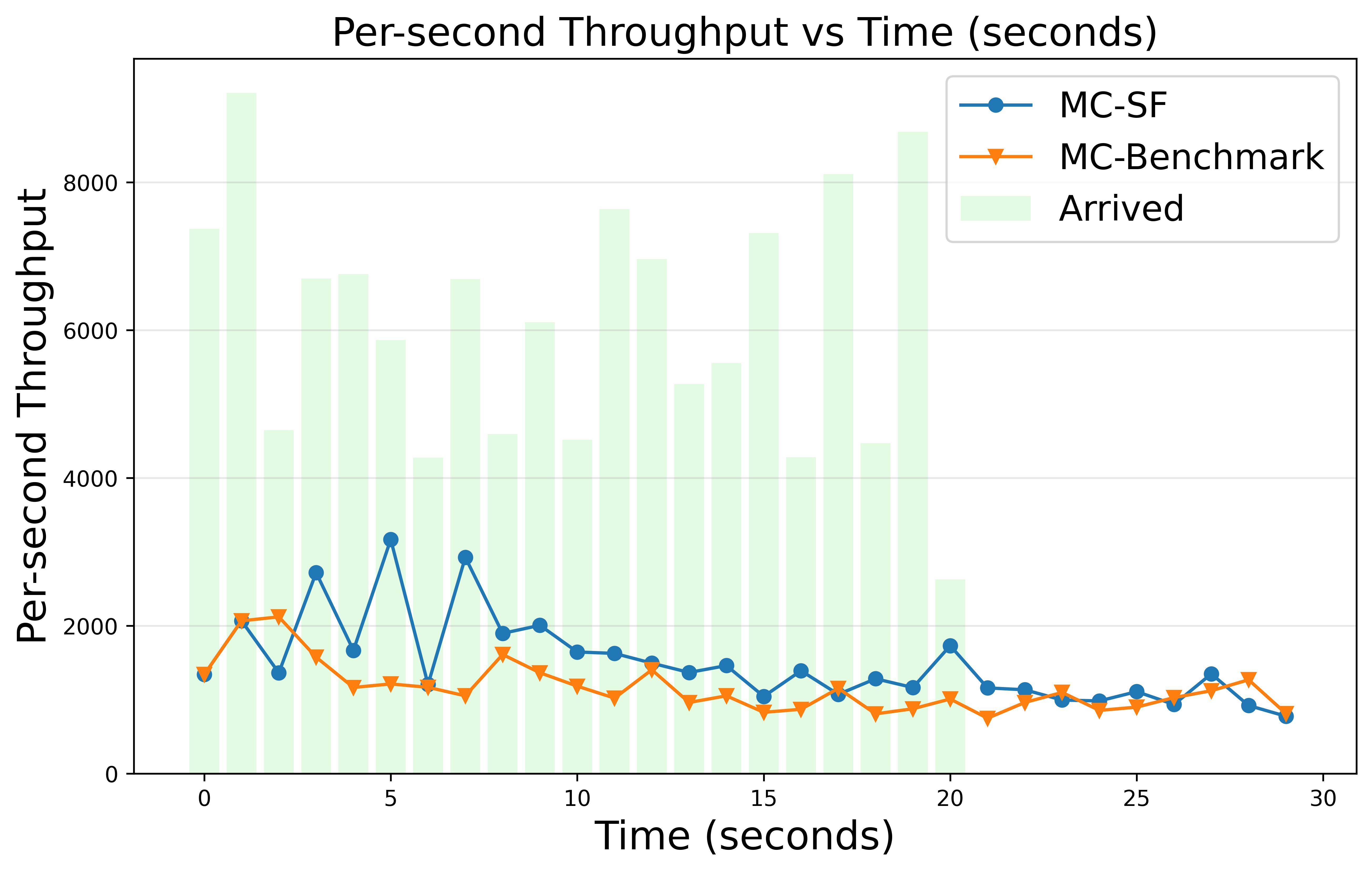}
\caption{Per-second Throughput Across Scheduling Algorithms.}
\label{fig:secondthroughput}
\end{figure}

Although our paper primarily targets latency minimization, we also report throughput as a secondary performance indicator. Figure~\ref{fig:secondthroughput} plots the instantaneous per-second throughput achieved by \mcsdf{} and \mcb{} for the first $1000$ arriving requests. The light green bars show the per-second arrival workload, measured as the total number of tokens introduced in each second (input$+$output per request). Under this overloaded regime, \mcsdf{} has higher processing throughput than \mcb{} for most time intervals, indicating that its latency improvements do not come at the expense of reduced service rate.

These results confirm that \mcsdf{} remains memory-safe and highly efficient across both regimes. Additional results are also provided in Appendix~\ref{app:num}.

\subsubsection{Performance under Prediction Errors}\label{subsubsec:pred_error}

In the previous experiments, we assumed that the scheduler has access to the true output length $o_i$ for each request $i$ (taken from the dataset). In practice, output length can be predicted and will inevitably contain error. This subsection evaluates the robustness of \mcsdf{} when only noisy length predictions are available.

\xhdr{Prediction Model.} We replace the true output length $o_i$ with a random prediction $\hat o_i$ generated as
\[
\hat o_i \sim \text{Uniform}\bigl((1-\epsilon)o_i,\,(1+\epsilon)o_i\bigr),
\]
where $\epsilon\in\{0.2,0.5,0.8\}$ controls the prediction error. The simulation setup (dataset, arrival process, inference-time estimation via Vidur, and latency metric) is identical to Section~\ref{subsec:realdata}, except that \mcsdf{} now uses $\hat o_i$ instead of $o_i$ when performing memory-feasibility checks and building batches.

\xhdr{Risk of Overflow and Memory Protection.} Because \mcsdf{} is designed to aggressively utilize available KV-cache memory, an underestimate $\hat o_i < o_i$ can cause the realized KV-cache growth to exceed the physical limit $M$. In our simulator, such an overflow triggers a clearing event, where all active requests are evicted and re-queued, increasing latency and potentially causing repeated retries.
To mitigate this underestimation risk, we introduce a simple protection margin mentioned above: we reserve a fixed fraction $\alpha M$ of memory and run \mcsdf{} as if the effective budget were $(1-\alpha)M$. Concretely, we set $\alpha=0.1$ and apply the same selection logic as before, but with $M$ replaced by $(1-\alpha)M$ in the feasibility check.

\begin{figure}[!tb]
\center
\includegraphics[width=0.6\textwidth]{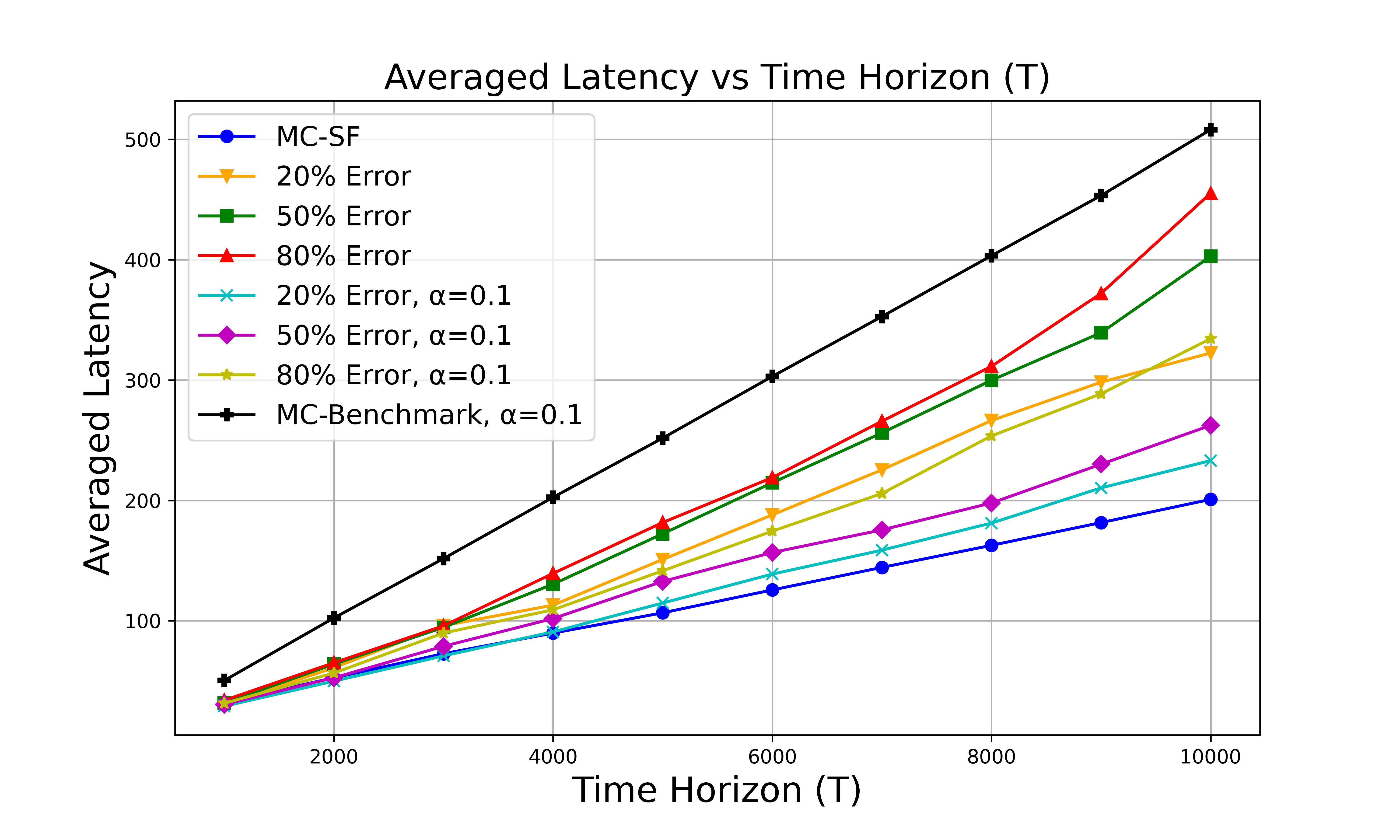}
\caption{Average End-to-End Latency Across Scheduling Algorithms Under Prediction Error.}
\label{fig:latencyerror}
\end{figure}

\xhdr{Results.} Figure~\ref{fig:latencyerror} reports the average end-to-end latency under different prediction error levels. As expected, larger prediction error leads to higher latency, since the scheduler must operate with noisier length estimates and a more conservative effective memory budget. Importantly, the $\alpha=0.1$ protection margin prevents the instability caused by systematic underestimation and substantially reduces the frequency and impact of clearing events. Even at $\epsilon=0.8$, \mcsdf{} with protection achieves significantly lower latency than the benchmark FCFS policy, highlighting that \mcsdf{} remains effective under substantial prediction noise.

%% file: conclusion.tex
\section{Discussion} \label{sec:conclusion}

{In this paper, we developed a formal model for online batching and scheduling in LLM inference that explicitly captures the dynamic memory growth induced by the KV cache. We introduced a hindsight-optimal benchmark via an integer programming formulation, established fundamental limits by showing that no deterministic online algorithm admits a constant competitive ratio under adversarial arrivals, and proposed a practical polynomial-time algorithm, \mcsdf{}, that achieves \emph{constant} competitive ratio under some structured conditions on the prompts that arrive. Through both synthetic experiments benchmarked against the hindsight optimal and large-scale simulations using real LLM inference traces, we demonstrated that \mcsdf{} achieves near-optimal latency and substantial improvements over existing scheduling heuristics while remaining memory-safe. Together, these results provide an end-to-end theoretical and empirical framework for understanding and designing scheduling policies for KV-cache-constrained LLM inference.}

There are several research directions that stem from our work. 

{Our model in Section~\ref{sec:model} assumes that upon arrival, each request is accompanied by a prediction $\tilde{o}_i$ that upper bounds its true output length $o_i$. An interesting future direction is to \emph{jointly} design prediction mechanisms for output lengths and batching–scheduling policies that operate simultaneously during inference. Ideally, such predictors would provide not only upper bounds on $o_i$, but also informative lower bounds, enabling tighter control of memory usage and more aggressive batching decisions. Our proposed algorithm and analysis can serve as a benchmark for this setting, characterizing the best achievable competitive performance when output lengths are either known exactly or upper bounded by a prediction algorithm. In fact, a recent follow-up work to our paper~\citep{chen2025adaptively} has taken one step closer to this direction by extending the model that we present in this work to a model where output lengths are predicted to be in a known interval. Finally, note that the literature on traditional scheduling has also provided online algorithms that only have ``predictions'' about the remaining time of jobs when they are being scheduled~\citep{azar2021flow,azar2022distortion,gupta2026optimal}.}

{Another important direction is to extend our framework to settings with \emph{multiple} computational workers operating in parallel. While the availability of multiple GPUs substantially enlarges the design space of batching and scheduling algorithms, it also introduces new challenges that are absent in the single-worker setting. For example, workers may be heterogeneous in terms of memory capacity or compute throughput, requiring ``matching'' decisions between requests and workers to maximize the overall efficiency of the entire system. Moreover, decisions must now balance load across workers while accounting for the evolving KV-cache footprint of each active request; see e.g., the recent work of \citet{balseiro2025load} on load balancing for LLM inference and the work of~\citet{jaiswal2025sageserve} for a more applied point-of-view on the subject. For settings where a prompt needs to be processed by multiple GPUs, it will be interesting to see how the literature on multi-server job scheduling (e.g.,~\citep{grosof2022optimal}) can provide insights for scheduling for LLM inference.}

{Finally, an important direction is to study other arrival models, even in the single-worker setting, where most requests are generated by an unknown stochastic process but a small fraction of requests are extreme outliers, being either unusually large or unusually small. Such heavy-tailed or mixed workloads are commonly observed in real LLM inference systems and pose challenges that are not well captured by purely adversarial or fully stochastic models. Our empirical results already suggest that algorithms such as \mcsdf{} remain effective under this type of workload heterogeneity. Moreover, our current theoretical analysis can serve as a benchmark for evaluating the performance of future algorithms designed for these hybrid stochastic–adversarial settings.}



%% file: Preliminary.tex
\newpage
\section*{\Large \begin{center}Supplementary Material for paper \\
\emph{``Online Scheduling for LLM Inference with KV Cache Constraints''}\end{center}}
\section{Background: Online Optimization in LLM Inference} \label{sec:pre}

This section provides background in LLM inference using a single computational worker (i.e., a single GPU).

\subsection{LLM Inference Process on a Single Request} \label{subsec:inferencesingle}

\begin{figure}[h]
\center
\includegraphics[width=0.64\textwidth]{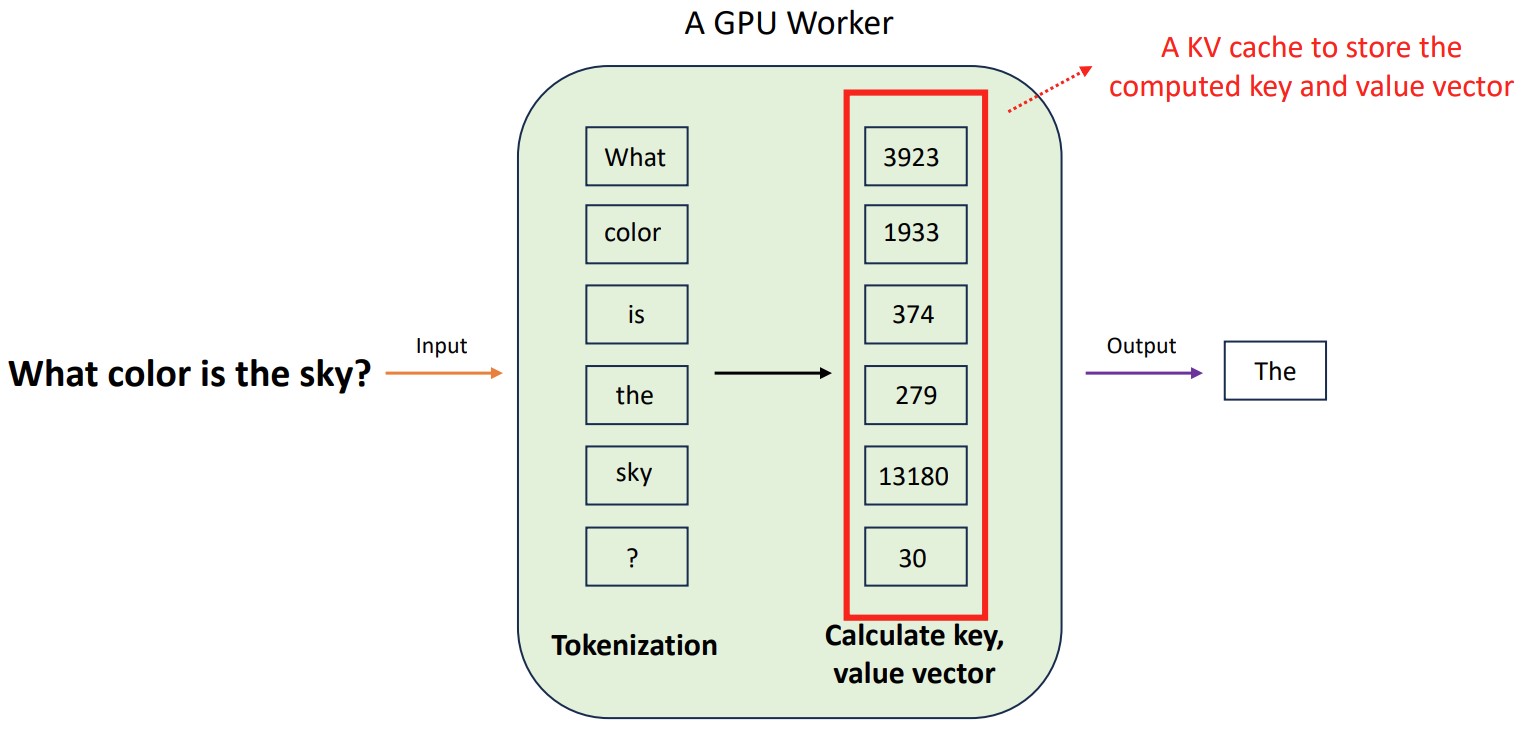}
\includegraphics[width=0.35\textwidth]{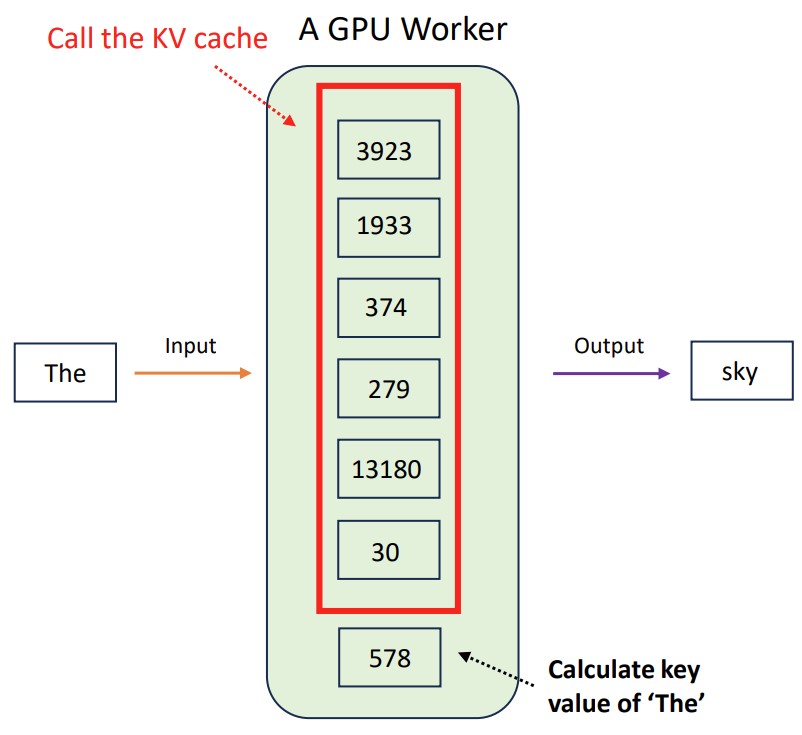}
\\
\includegraphics[width=0.4\textwidth]{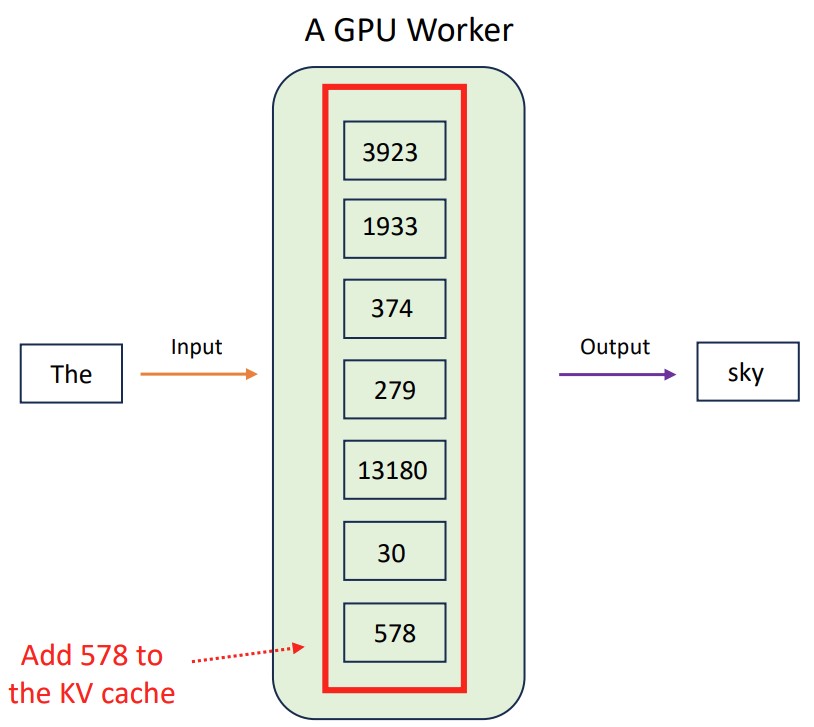}
\text{ } \text{ }
\includegraphics[width=0.4\textwidth]{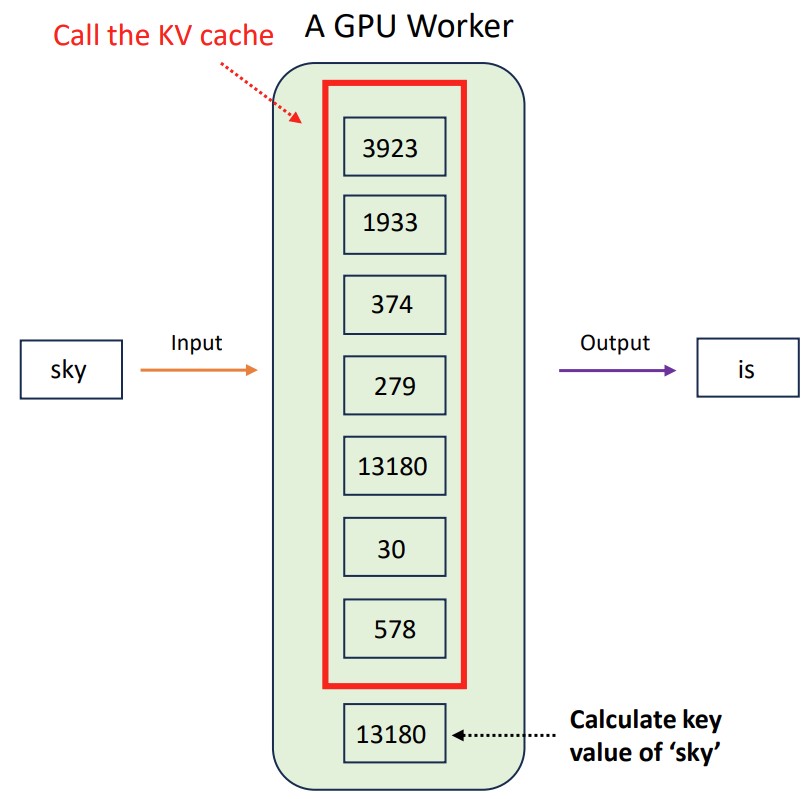}
\caption{An example of LLM inference \citep{llminference}. } 
\label{fig:LLMexp1}
\end{figure}

We first demonstrate how a single GPU worker processes an incoming prompt, using the example prompt “What color is the sky?” from \cite{llminference}. The workflow is illustrated in Figure \ref{fig:LLMexp1}. 

Upon receiving the prompt, as shown in the top left figure, the worker begins by tokenizing the prompt and generates key and value vectors. Then, it uses these vectors to calculate the attention scores, which indicate the relevance of each token, enabling the generation of the first output token, “The.” To avoid recalculating these vectors with each new token, the worker saves them in a key-value (KV) cache.

Next, as depicted in the top right figure, after generating “The,” the worker processes this token by calculating its key and value and retrieving the key-value vectors of the initial prompt tokens from the KV cache. Using this combined data, the model generates the next token, “sky.”

The bottom left figure shows that during “sky” generation, the worker calculates the key and value for “The” and adds them to the KV cache to avoid recalculations. Finally, as illustrated in the bottom right figure, the worker processes “sky,” continuing this cycle until all output tokens are generated.

Finally, we note that when a prompt request arrives, the computational worker does not know the length of its response (i.e., the number of tokens it will generate). However, certain techniques can predict the output length of each prompt before processing begins. For instance, \cite{zheng2024response} proposes a methodology that achieves an output length prediction accuracy of up to 81\%.

\subsection{Illustration of Novel Difficult Aspects of LLM Inference Scheduling} \label{subsec:diff}

Though scheduling problems have been studied extensively in the previous literature on computing resource allocation (e.g., \cite{pinedo2012scheduling}), we illustrate some of the new dynamics of LLM inference that separates it from standard scheduling problems. 
To be specific, due to the autoregressive LLM inference process described in \Cref{subsec:inferencesingle}, the memory size required by a request keeps growing during the inference procedure, which is fundamentally different from the previous scheduling models where the size (or the resource occupation) of a job is fixed. 

To see this point, suppose that we are following the principle of processing the shortest job first and there are two prompts P1 and P2, whose sizes eventually grow to $t_1$ and $t_2$. If $t_1+t_2>M$, where $M$ is the memory limit of the KV cache, then the two prompts cannot be processed at the same time in the classical scheduling model. However, this may not be the case for our LLM inference model since the sizes are varying over time. Suppose that the initial size for both prompts is the same, denoted by $s$, and it satisfies that $s\leq t_1 \leq M/2$. The size of both jobs increases by $1$ at each round until reaching $t_1$ and $t_2$. Then, we can process P1 and P2 at the same time until the time $t_1-s$ when the size of both jobs increases to $t_1$; note that we can process both P1 and P2 concurrently, since the memory requirement for the two jobs together is upper bounded $2t_1\leq M$. After time $t_1-s$, though the size of P2 may be larger than $t_1$, prompt P1 has finished processing (since we have processed $s + t_1 - s = t_1$ tokens), and the required memory can be released to further process P2. Therefore, P1 and P2 can be processed at the same time in the LLM inference model. 

Moreover, the batching problem in our setting significantly differs from traditional problems (e.g., \cite{potts2000scheduling}). When assuming that the KV cache memory limit is the sole constraint for the batch size, the number of jobs in a batch during each round varies depending on the specific jobs included in the batch. This is because the memory usage for each output token increases linearly over time. While some studies (\cite{kashan2008scheduling,hazir2014batch,yang2022scheduling}) consider batch size constraints as the total weight of jobs—where weight corresponds to the memory usage of each prompt or token—in our problem, the situation is more complex. Processing a prompt or token increases the memory usage. Consequently, the effective weight of jobs at any given time depends on the starting time to process this token, and the index of this token in the request. This weight dependency also makes the problem more challenging.

%% file: Append1.tex
\section{Appendix for Section \ref{sec:stochastic}} \label{app:sto}

\subsection{Proof of Theorem \ref{thm:advr}}

\begin{proof}[Proof of Theorem \ref{thm:advr}]
    Fix a deterministic algorithm $\cA$. For any $M \ge 1$, consider the following instance $\cI$ with available memory $M$ and with $s_i=1$ for all request $i$. First, a request is released at time 0 with output length $o_1 = M-1$ (``long request''). Let $b$ be the time $\cA$ starts processing this first request. Then $\frac{M}{2}$ requests are released at time $r := b + M - \frac{\sqrt{M}}{2}$ with output size $o_i = 1$ (``short requests''). Let $n := \frac{M}{2} + 1$ denote the number of requests in this instance. We claim that the total latency of $\cA$ on this instance is at least $\Omega(\sqrt{n}) \cdot \OPT(\cI)$. 
    
    For that, we first claim that 
    \begin{align}
    \OPT(\cI) \le 3.5M. \label{eq:UBOPT}
    \end{align}
   If $r \ge M$, then it is possible to start processing the long request at time 0 and finish by time $M-1$ (incurring latency $M-1$) and then at time $r \ge M$ start processing all the short requests (since the maximum memory occupation of each short request is 2, they can all be executed together), finishing them at time $r + 1$ (incurring latency 1 for each of these requests); in this case we have $\OPT(\cI) \le M-1 + \frac{M}{2} \le 3.5M$ as desired.

    If $r < M$, then a possible offline solution is to first do all the short requests and then the long request, namely start executing all the short requests at time $r$, finishing them at time $r + 1$ (incurring latency $1$ for each), and then start to process the long request at time $r + 2$, finishing it at time $r + 2 + (M-1)$ (incurring latency $r + 2 + (M-1)$). Therefore, $\OPT(\cI) \le \frac{M}{2} + r + 1 + M \le 2.5 M + 1 \le 3.5 M$, where the second inequality uses the fact we are in the case $r < M$ and the last inequality uses $M \ge 1$. Thus, in both cases we have that \eqref{eq:UBOPT} holds.  

    We now lower bound the latency that $\cA$ experiences over $\cI$. In all times between time $r = b + (M-1) - \frac{\sqrt{M}}{2}$ and $b + (M-1)$, the long request occupies memory at least $M - \frac{\sqrt{M}}{2}$; thus, the total memory available for other requests across all those times is at most $\frac{\sqrt{M}}{2} \cdot \frac{\sqrt{M}}{2} = \frac{M}{4}$. Thus, at least $\frac{M}{2} - \frac{M}{4} = \frac{M}{4}$ of the short requests can only be started by the algorithm $\cA$ on or after time $b + M$, and thus each of them incurs latency at least $b + M - r = \frac{\sqrt{M}}{2}$. Thus, the total latency of $\cA$ is at least $\frac{M}{4} \cdot \frac{\sqrt{M}}{2}$. 
    
    Combining this with the upper bound on $\OPT$ from \eqref{eq:UBOPT}, we get
    \begin{align*}
 \frac{ \text{TEL}(\cI; \mathcal{A})}{\text{OPT}(\mathcal{I})} \ge \frac{\sqrt{M}}{28}.
    \end{align*}
    Since the number of items in the instance is $n = 1 + \frac{M}{2}$, this shows that the competitive ratio of $\cA$ cannot be better than $\Omega(\sqrt{n})$, thus proving the theorem.      
\end{proof}


\subsection{Proof of Proposition~\ref{thm:complexity}}

\begin{proof}[Proof of Proposition~\ref{thm:complexity}]
At each time step $t \in [T]$, to solve Eq.~\eqref{eqn:Selection}, we add requests in ascending order of their indices, stopping the process once any inequality is violated. The memory limit $M$ is a constant, which implies that we will add at most $M/(\max_{i \in [n]}s_i+1)=O(M)$ requests in $U^{(t)}$. For each request $i$, to decide whether to add it or not in $U^{(t)}$, one has to check all inequalities at the completion times $p_j+o_j$ for $j \in S^{(t)} \cup U^{(t)}$ in Eq.~\eqref{eqn:Constraint}. The number of inequalities we need to check is $O(M)$ since $|S^{(t)} \cup U^{(t)}|=O(M)$. Therefore, the feasibility check for each request $i$ has a complexity of $O(M)$. Since we can add at most $O(M)$ number of requests to a batch, the complexity at time $t$ is at most $O(M^2)$.
\end{proof}

%% file: robustUBApp.tex
\subsection{Proof of Theorem \ref{thm:main} with Inaccurate Output Length Predictions} \label{app:stochastic}

Recall that in Section \ref{sec:stochastic} we proved that the algorithm $\alg$ is $O(1)$-competitive when the output length predictions $\po_i$ are exact, i.e., $\po_i o_i$ for all $i$. Now we show the required modifications in the proof to accommodate approximate predictions, namely when $o_i \le \po_i \le \alpha o_i$ for all requests $i$ for some constant $\alpha \ge 1$; this will provide a full proof of Theorem \ref{thm:main}.

\paragraph{Upper bound on the total latency of $\alg$.}

First, we define the modified quantity $\tilde{n}_o$ as the number of requests in the instance whose \emph{predicted} output $\po_i$ equals $o$. We analogously define the modified requests groups $\tilde{U}_\ell$, namely for $\ell = 0,\ldots, \lfloor \log \po_{\max} \rfloor$ ($\po_{\max}$ is the largest $\po_i$), let $\tilde{U}_\ell$ denote the set of requests that have predicted output length in the interval $[2^\ell, 2^{\ell + 1})$. 

In the presence of imperfect predictions, Lemma \ref{lemma:UBo2} becomes the following.

\begin{lemma} \label{lemma:UBo2}
	Consider one of the sets of requests $\tilde{U}_\ell$, and let $\underline{o}_\ell := 2^{\ell}$ and $\bar{o}_\ell := 2^{\ell + 1} - 1$ denote the smallest and largest predicted possible output length for requests in this set. Let $\underline{t}$ and $\bar{t}$ be the first and last time the algorithm $\alg$ processes a request in $\tilde{U}_{\ell}$. Then the distance between these times can be upper bounded as 
	$$\bar{t} - \underline{t} \,\le\, \frac{192\alpha^2}{M} \cdot \sum_{\po = \underline{\po}_\ell}^{\bar{o}_\ell} \tilde{n}_o \cdot \vol_o + 5 \bar{o}_\ell.$$
\end{lemma}

\begin{proof}
(For the remainder of the proof we omit the subscript $\ell$ in $\underline{o}_\ell$ and $\bar{o}_\ell$.) The proof follows the same steps as that of the old Lemma \ref{lemma:UBo2}. We start by partitioning the interval $\{\underline{t},\ldots,\bar{t}\}$ into disjoint subintervals $I_1,I_2,\ldots,I_w$ of length $\bar{o}$ (where $I_w$ is the only exceptional interval that can be smaller than $\bar{o}$). Again, for an interval $I \subseteq [T]$, let $\peak(I)$ be the \emph{actual} (not predicted) peak memory use by $\alg$ during this interval. Slightly abusing notation, also let $\vol_\ell(I)$ be the total actual amount of memory that the requests $\tilde{U}_\ell$ occupy in $\alg$'s schedule added up over all times in the interval $I$. The next claim is the ``peak-to-volume'' Claim \ref{claim:vol-around} in the presence of uncertain predictions. 

\begin{claim}[Peak-to-volume] \label{claim:vol-around2}
	Consider any 3 consecutive intervals $I_j, I_{j+1}, I_{j+2}$ of length $\bar{o}$. Then $\vol_\ell(I_j \cup I_{j+1} \cup I_{j+2})\ge \frac{1}{4\alpha^2}\, \peak(I_{j+1}) \cdot \frac{\vol_{\bar{o}}}{s + \bar{o}}$.
\end{claim}

\begin{proof}[Proof of Claim~\ref{claim:vol-around2}]
	Let $\tilde{t} \in I_{j+1}$ be the time when the peak memory occupation $\peak(I_{j+1})$ happens. Since the interval $I_{j+1}$ is strictly between times $\underline{t}$ and $\bar{t}$, by definition of the algorithm it only processes requests of $\tilde{U}_\ell$, and so only those contribute to the memory occupation at time $\tilde{t}$. If $k$ of these requests contribute to this peak occupation, then each contributes at most $s + \bar{o}$ to it (recall that the true output length is always at most the predicted output length, which is at most $\bar{o}$ for requests in $\tilde{U}_\ell$); thus, we have $k \ge \frac{\peak(I_{j+1})}{s+\bar{o}}$. Since such request is completely processed in the bigger interval $I_j \cup I_{j+1}\cup I_{j+2}$, each such request contributes with at least $\vol_{\underline{o}/\alpha}$ to the memory volume $\vol_\ell(I_j \cup I_{j+1}\cup I_{j+2})$ (recall that the true output length is always at least $\frac{1}{\alpha}$ of the predicted output length); hence, $\vol_\ell(I_j \cup I_{j+1}\cup I_{j+2}) \ge k \cdot \vol_{\underline{o}/\alpha}$. Finally, since $\underline{o}/\alpha \ge \frac{1}{2\alpha} \bar{o}$, a quick calculation shows that $\vol_{\underline{o}/\alpha} \ge \frac{1}{4\alpha^2} \vol_{\bar{o}}$. Combining these threes inequalities gives the claim. 
\end{proof}

	We now conclude the proof of Lemma \ref{lemma:UBo2}.	Suppose for contradiction that $\bar{t} - \underline{t} > \frac{192\alpha^3}{M} \sum_{o = \underline{o}}^{\bar{o}} \tilde{n}_o \cdot \vol_o + 5\bar{o}$. First, since $\bar{o} \le 2 \underline{o}$, again we have  $\vol_o \ge \frac{1}{4} \vol_{\overline{o}}$ for all $o$ between $\underline{o}$ and $\bar{o}$. Then since $|\tilde{U}_\ell| = \sum_{o = \underline{o}}^{\bar{o}} \tilde{n}_o$, our assumption implies that $\bar{t} - \underline{t} > \frac{48\alpha^3}{M} \cdot |\tilde{U}_\ell| \cdot \vol_{\bar{o}} + 5\bar{o}$. 
	
	This further implies that $w$ (the number of the intervals $I_j$ of length $\bar{o}$ in this period) is at least 
	\begin{align*}
	w \ge \bigg\lfloor \frac{\frac{48\alpha^3}{M} \cdot |\tilde{U}_\ell| \cdot \vol_{\bar{o}} + 5\bar{o}}{\bar{o}}  \bigg\rfloor \ge \frac{48\alpha^3}{M} \cdot |\tilde{U}_\ell| \cdot \frac{\vol_{\bar{o}}}{\bar{o}} + 4.
	\end{align*}
	Every such interval $I_j$ other than $I_w$ has real peak memory utilization 
    more than $\frac{1}{\alpha} (M - (s+\bar{o}))$: otherwise even using the predicted output lengths $\po_i \le \alpha o_i$, which is what effectively the memory constraint check in $\alg$ uses, there would be a time in this interval where the algorithm would have scheduled an additional request from $\tilde{U}_\ell$.
    Then applying Claim~\ref{claim:vol-around} to the first $\lfloor \frac{w-1}{3} \rfloor$ groups of 3 consecutive intervals $I_j$'s, we obtain that 
	\begin{align}
	\vol_\ell(\{\underline{t},\ldots,\bar{t}\}) &\ge \bigg\lfloor \frac{w-1}{3} \bigg\rfloor \cdot \frac{1}{4\alpha^2} \cdot \frac{(M - (s+\bar{o}))}{\alpha} \cdot \frac{\vol_{\bar{o}}}{s+\bar{o}} \nonumber \\
	& \ge \frac{4}{M} \cdot |\tilde{U}_\ell| \cdot \frac{\vol_{\bar{o}}}{\bar{o}} \cdot \frac{M}{2} \cdot \frac{\vol_{\bar{o}}}{s+\bar{o}} > |\tilde{U}_\ell| \cdot \vol_{\bar{o}}, \label{eq:contradiction2}
	\end{align}
	where the second inequality uses the assumption $M$ is twice as big as the maximum single request predicted occupation, i.e., $s + \bar{o} \le \frac{M}{2}$, and the last inequality uses $\vol_{\bar{o}} = s \cdot \bar{o} + \frac{\bar{o} \cdot (\bar{o}+1)}{2} > \frac{1}{2} \bar{o} \cdot (s + \bar{o})$. 
	
	However, each request $\tilde{U}_\ell$ contributes at most $\vol_{\bar{o}}$ to $\vol_\ell(\{\underline{t},\ldots,\bar{t}\})$, and thus $\vol_\ell(\{\underline{t},\ldots,\bar{t}\}) \le |\tilde{U}_\ell| \cdot \vol_{\bar{o}}$; this contradicts Equation~\eqref{eq:contradiction2}, and concludes the proof of the lemma. 	
\end{proof}

    We then get the following upper bound on $\alg$ that generalizes Lemma \ref{lemma:UBAlg}.

\begin{lemma}[UB on $\alg$] \label{lemma:UBAlg2}
	The total latency incurred by the algorithm $\alg$ is at most
	\begin{align*}
		\frac{1536\alpha^3}{M} \sum_{o} \tilde{n}_{o} \cdot \sum_{o' \le o} \tilde{n}_{o'} \cdot  \vol_{o'} + 24 \sum_{o} \tilde{n}_{o} \cdot o
	\end{align*}
\end{lemma}

\begin{proof}
	Let $t_\ell$ be the first time a request in $\tilde{U}_\ell$ is (starting to be) processed by $\alg$ (let $\ell_{\max} := \lfloor \log \po_{\max} \rfloor$ and let $t_{\ell_{\max} + 1}$ be the last time the algorithm is processing something). The latency for each request in $\tilde{U}_\ell$ is at most {$t_{\ell+1} + \bar{o}_\ell$} (i.e., if the algorithm starts processing requests from the next group then is has already started to process all requests in $\tilde{U}_\ell$, which take at most $+ \bar{o}_\ell$ time to complete); thus, the total latency is at most $\sum_\ell |\tilde{U}_\ell| \cdot ({t_{\ell+1} + \bar{o}_\ell})$. However, from Lemma~\ref{lemma:UBo2} we know that $$t_{\ell+1} - t_{\ell} \,\le\, \frac{192\alpha^3}{M} \cdot |\tilde{U}_\ell| \cdot \vol_{\bar{o}_\ell} + 5\bar{o}_\ell + 1 \,\le\, \frac{192\alpha^3}{M} \cdot |\tilde{U}_\ell| \cdot \vol_{\bar{o}_\ell} + 6\bar{o}_\ell,$$ and so $t_{\ell + 1} \le \frac{192\alpha^3}{M} \sum_{\ell' \le \ell} |\tilde{U}_{\ell'}| \cdot \vol_{\bar{o}_{\ell'}} + 6 \sum_{\ell' \le \ell} \bar{o}_{\ell'}$.    This gives that the total latency of $\alg$ can be upper bounded as 
	\begin{align}
		\alg \,\le\, \frac{192\alpha^3}{M} \underbrace{\sum_\ell |\tilde{U}_\ell| \sum_{\ell' \le \ell} |\tilde{U}_{\ell'}| \cdot \vol_{\bar{o}_{\ell'}}}_{A} + 6 \underbrace{\sum_\ell |\tilde{U}_\ell| \sum_{\ell' \le \ell} \bar{o}_{\ell'}}_{B}.  \label{eq:UBAlg2}
	\end{align}
	Concluding the proof of the lemma requires just a bit of algebra to clean up the bound. 
	
	To upper bound the term $A$, let $\tilde{O}_\ell := \{\underline{o}_\ell, \ldots, \bar{o}_\ell\}$ be the possible predicted output lengths of the requests in $\tilde{U}_\ell$. Again we observe that since $\bar{o}_{\ell} \le 2\underline{o}_\ell$, we have $\vol_{\bar{o}} \le 4 \vol_o$ for every $o \in \tilde{O}_\ell$. Moreover, since $|\tilde{U}_\ell| = \sum_{o \in \tilde{O}_\ell} \tilde{n}_o$, we have 
	\begin{gather*}
	|\tilde{U}_\ell|^2 \cdot \vol_{\bar{o}_\ell} \le 2 \bigg(\sum_{o \in \tilde{O}_\ell} \tilde{n}_o \sum_{o' \in \tilde{O}_\ell, o' \le o} \tilde{n}_{o'}\bigg) \cdot  \vol_{\bar{o}_\ell} \le  8 \sum_{o \in \tilde{O}_\ell} \tilde{n}_o \sum_{o' \in \tilde{O}_\ell, o' \le o} \tilde{n}_{o'} \cdot  \vol_{o'}
	\end{gather*}
	and for $\ell' < \ell$ 
	\begin{gather*}
	|\tilde{U}_\ell| \cdot |\tilde{U}_{\ell'}| \cdot \vol_{\bar{o}_{\ell'}} \le \bigg(\sum_{o \in \tilde{O}_\ell} \tilde{n}_o \sum_{o' \in \tilde{O}_{\ell'}} n_{o'}\bigg) \cdot  \vol_{\bar{o}_{\ell'}} \le  4 \sum_{o \in \tilde{O}_\ell} \tilde{n}_o \sum_{o' \in \tilde{O}_{\ell'}} \tilde{n}_{o'} \cdot  \vol_{o'},
	\end{gather*}	
	which combined give
	\begin{gather*}
	|\tilde{U}_\ell| \sum_{\ell' \le \ell} |\tilde{U}_{\ell'}| \cdot \vol_{\bar{o}_{\ell'}} \,\le\, 8  \sum_{o \in \tilde{O}_\ell} \tilde{n}_o \sum_{o' \le o} \tilde{n}_{o'} \cdot  \vol_{o'},
	\end{gather*}	
	and so adding up over all $\ell$ gives the upper bound $A \le 8 \sum_o \tilde{n}_o \sum_{o' \le o} \tilde{n}_{o'} \cdot  \vol_{o'}.$
	
	To upper bound the term $B$ in \eqref{eq:UBAlg2}, we observe that since the $\bar{o}_\ell$'s grow exponentially, $\sum_{\ell' \le \ell} \bar{o}_{\ell'} \le 2 \bar{o}_{\ell}.$ Then since $\bar{o}_{\ell} \le 2 o$ for every $o \in \tilde{O}_{\ell}$, we have
	\begin{align*}
		B \le 2 \sum_\ell |\tilde{U}_\ell| \cdot \bar{o}_{\ell} = 2 \sum_\ell \sum_{o \in \tilde{O}_\ell} \tilde{n}_o \cdot \bar{o}_{\ell} \le  4 \sum_\ell \sum_{o \in \tilde{O}_\ell} \tilde{n}_o \cdot o = 4 \sum_o \tilde{n}_o \cdot o.
	\end{align*}
	Plugging these upper bounds on $A$ and $B$ on \eqref{eq:UBAlg2}, we get
	\begin{align*}
		\alg \le \frac{1536\alpha^3}{M} \sum_o \tilde{n}_o \sum_{o' \le o} \tilde{n}_{o'} \cdot  \vol_{o'} + 24 \sum_o \tilde{n}_o \cdot o.
	\end{align*}
	This concludes the proof of Lemma \ref{lemma:UBAlg2}.
\end{proof}


\paragraph{Lower bound on the total latency of $\OPT$.} 

In the presence of incorrect predictions, we have the following lower bound on the optimal latency, which mirrors Lemma \ref{lemma:LBOPT}.

\begin{lemma} \label{lemma:LBOPT2}
	The total optimal latency $\OPT$ satisfies:
	\begin{align*}
		\OPT \ge \frac{1}{6M\alpha^2} \sum_o \tilde{n}_o \cdot \sum_{o' \le o} \tilde{n}_{o'} \cdot \vol_{o'} + \frac{1}{6\alpha} \sum_o \tilde{n}_o \cdot o.
	\end{align*}
\end{lemma}

For this, we consider the same LP relaxation as before, with the required adjustments. Namely, let $\tilde{U}_o$ denote the set of requests with predicted output length $o$ (so $|\tilde{U}_o| = \tilde{n}_o$); thus the true volume of each such request is at least $\vol_{o/\alpha} \ge \frac{\vol_o}{\alpha^2}$. Let $\bar{a}^t_o$ be the number of requests in $\tilde{U}_o$ that finish at time $t$ in the optimal solution. The memory volume of all requests that finish up to time $t$ need to fit in the total memory $t \cdot M$ available up to that time, and hence $\sum_{t' \le t} \sum_o \bar{a}^t_o \cdot \frac{\vol_o}{\alpha^2} \le t \cdot M$. Moreover, $\sum_t \bar{a}^t_o = \tilde{n}_o$ (all requests in $\tilde{U}_o$ finish at some time). Finally, the $\sum_o \bar{a}^t_o$ requests that finish at time $t$ have latency (recall all requests are released at time 0) equal to $t$, and the optimal latency $\OPT$ is given by $\sum_t t \cdot \sum_o \bar{a}^t_o$. Together these observations show that $\OPT$ can be lower bounded by the following Linear Program with variables $a^t_o$, where in particular we relax the requirement that $\bar{a}^t_o$'s are integers:
\begin{align}
	\OPT_{LP} := \min &\sum_t t \cdot \sum_o a_o^t \notag\\
	\textrm{s.t.}\,& \sum_{t' \le t} \sum_o a^t_o \cdot \vol_o \le t \cdot M \alpha^2, \qquad \forall t \label{eq:LPmem2}\\
	&\sum_t a^t_o = \tilde{n}_o, \qquad \forall o \notag\\
	&a^t_o \ge 0, \qquad \forall t,o. \notag
\end{align}

We then lower bound $\OPT_{LP}$. Consider the optimal solution $\{a^{* t}_o\}_{t,o}$ for this LP. For a given output size $o$, let $t^*_o$ be the first time $t$ where $a^{* t}_o > 0$, i.e., where a request $U_o$ is assigned to time $t$. Using exactly the same argument as in the proof of Lemma \ref{lemma:LBOPT} we have that the ``first time'' values $t^*_o$ are non-decreasing, namely $t^*_o \le t^*_{o'}$ when $o < o'$. 

Let $t^*_{o_{\max} + 1}$ be the last time such that $\sum_o a^{* t}_o > 0$. Using this observation we will prove the following lower bound on the ``first times'' $t^*_o$.

\begin{claim}	\label{claim:tStar}
	For all $o$ we have $t^*_{o + 1} \ge \frac{1}{M\alpha^2} \sum_{o' \le o} \tilde{n}_{o'} \cdot \vol_{o'}$. 
\end{claim}

\begin{proof}
	By the observation above, in the optimal solution $\{a^{*o'}_t\}_{o',t}$ all items with output length at most $o$ are assigned to times $\le t^*_{o + 1}$, i.e., no later than when the next output length is assigned; thus, $\sum_{t' \le t^*_{o + 1}} a^{*t'}_{o'} = \tilde{n}_{o'}$ for all $o' \le o$. Then considering \eqref{eq:LPmem2} to time $t^*_{o + 1}$ we get
	\begin{align*}
		\sum_{o' \le o} \tilde{n}_{o'} \cdot \vol_{o'} = \sum_{t' \le t^*_{o + 1}} \sum_{o' \le o} a^{*t'}_{o'} \cdot \vol_{o'} \le t^*_{o + 1} \cdot M\alpha^2,
	\end{align*}
	and rearranging we get the claim. 
\end{proof}

We are now able to prove the lower bound on $\OPT$ from Lemma \ref{lemma:LBOPT2}.

\begin{proof}[Proof of Lemma \ref{lemma:LBOPT2}]
	By definition of $t^*_o$, we know that $a^{*t}_o = 0$ for all $t < t^*_o$, and so $\sum_t t \cdot a^{*t}_o \ge t^*_o \cdot \sum_{t \ge t^*_o} a^{*t}_o = t^*_o \cdot \tilde{n}_o$. Plugging the bound on $t^*_o$ from the previous claim and adding over all $o$ we can lower bound $\OPT_{LP}$ (and thus $\OPT$) as
	\begin{align}
	\OPT \ge \OPT_{LP} &= \sum_t t \cdot \sum_o a^{*t}_o \notag\\
	&\ge \sum_o \tilde{n}_o \cdot t^*_o \notag\\
	&\ge \sum_o \tilde{n}_o \cdot \bigg( \frac{1}{M\alpha^2} \sum_{o' < o} \tilde{n}_{o'} \cdot \vol_{o'} \bigg) \notag\\
	&=  \frac{1}{M\alpha^2} \sum_o \tilde{n}_o \sum_{o' \le o} \tilde{n}_{o'} \cdot \vol_{o'} - \frac{1}{M\alpha^2} \sum_o \tilde{n}_o^2 \cdot \vol_o.\label{eq:LBOPT12}
	\end{align}
	To remove the negative term in the right-hand side (and add a new one, to match that of Lemma \ref{lemma:UBAlg2}), we provide two other lower bounds on the original $\OPT$ (not the LP). 
	
	The first is that for any predicted output length $o$, due to memory constraints, at most $\frac{\tilde{n}_o}{2}$ of them can be finished in the optimal schedule by time $\frac{\tilde{n}_o}{2} \frac{\vol_o}{M\alpha^2}$, since the total memory available up to this time is $\frac{\tilde{n}_o}{2} \cdot \frac{\vol_o}{\alpha^2}$ and each such request consumes at least $\frac{\vol_o}{\alpha^2}$ memory; thus, the total latency on the optimal solution for the request of predicted output length $o$ is at least $\frac{\tilde{n}_o}{2} \cdot \frac{\tilde{n}_o}{2} \frac{\vol_o}{M\alpha^2}$. Adding this over all $o$, the total latency $\OPT$ has the lower bound 
	\begin{align}
	\OPT \ge \frac{1}{4M\alpha^2} \sum_o \tilde{n}^2_o \cdot \vol_o.  \label{eq:LBOPT22}
	\end{align}	
	For the second additional lower bound, since each request of predicted output length $o$ takes at least $\frac{o}{\alpha}+1$ units of processing/time to finish (and thus has latency at least $\frac{o}{\alpha}$), we also have $\OPT \ge \sum_o \tilde{n}_o \cdot \frac{o}{\alpha}$.
	
	Adding this bound plus 4 times \eqref{eq:LBOPT22} plus \eqref{eq:LBOPT12} we get 
	\begin{align}
	6\, \OPT \ge  \frac{1}{M\alpha^2} \sum_o \tilde{n}_o \sum_{o' \le o} \tilde{n}_{o'} \cdot \vol_{o'} + \sum_o \tilde{n}_o \cdot \frac{o}{\alpha}.
	\end{align}	 
	This concludes the proof of Lemma \ref{lemma:LBOPT2}.
\end{proof}


Combining the bound on $\alg$ from Lemma \ref{lemma:UBAlg2} and the lower bound on $\OPT$ from Lemma \ref{lemma:LBOPT2}, we obtain $\alg \le \alpha^5 \cdot O(1) \cdot \OPT$, which is $O(1) \cdot \OPT$ as $\alpha$ is taken to be constant. Thus, this concludes the proof of Theorem \ref{thm:main}.

%% file: Append4.tex
\section{Appendix for Section \ref{sec:num}} \label{app:num}

All experiments were conducted on a Microsoft Surface Laptop with Snapdragon® X Elite (12 Core) processor.



\begin{algorithm}[htbp] \renewcommand{\algorithmcfname}{Algorithm}
    \SetAlgoLined
    \caption{Memory Constrained Benchmark (\mcb)}
    \label{alg:2}
\DontPrintSemicolon
\SetAlgoLined
\KwIn{Memory capacity $M$, time horizon $T$}
\KwOut{Schedule for processing requests}
\For{each round $t = 1$ to $T$}{

    Let $S^{(t)}$ be the set of requests that have already stated processing, and let $R^{(t)}$ be the set of waiting requests at time $t$. Also set $U^{(t)} = \emptyset$\;
    
    \For{each request $i \in R^{(t)}$ in ascending order of {arrival time $a_i$}}{
        Set a time list $t' = p_j + o_j$ for $j \in S^{(t)}  \cup U^{(t)} \cup \{i\} $\;
        
        \If{all inequalities in Equation~\eqref{eqn:Constraint} hold for all $t'$}{
            Add request $i$ to $U^{(t)}$\;
        }
        \Else{
            Break the for loop\;
        }
    }
    
    Process the requests in $S^{(t)}  \cup U^{(t)}$\;
    
}
\end{algorithm}

In this section, we provide additional details for the numerical experiments with real data.  
We take the memory capacity as 16,492 as observed in real experiments reported to us by systems engineers through private communication.

\begin{figure}[!tb]
\center
\includegraphics[width=0.8\textwidth]{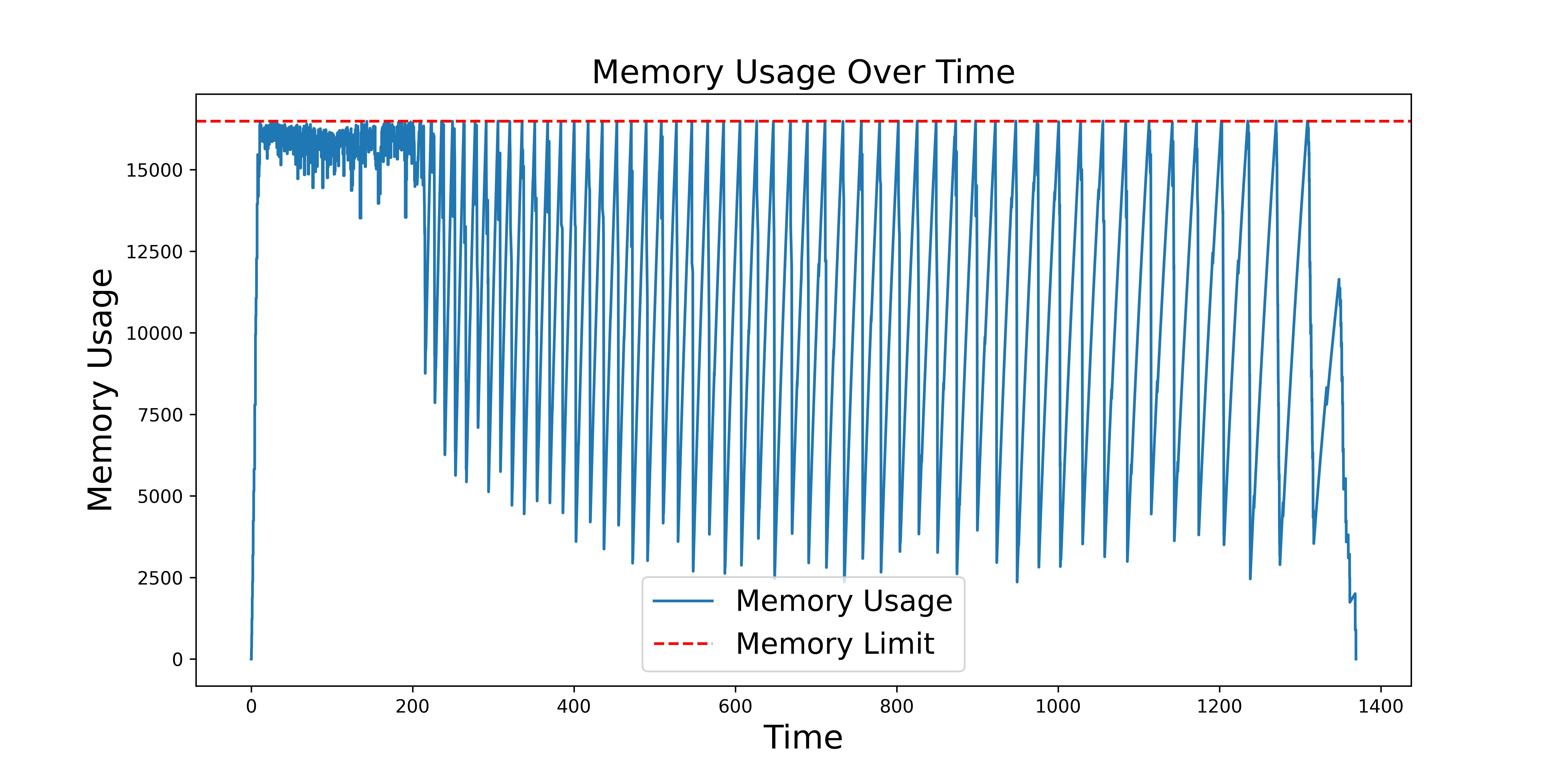}
\caption{Memory Usage over Time for Algorithm~\ref{alg:1} in the High Demand Case} 
\label{fig:movertime}
\end{figure}

\begin{figure}[!tb]
\center
\includegraphics[width=0.45\textwidth]{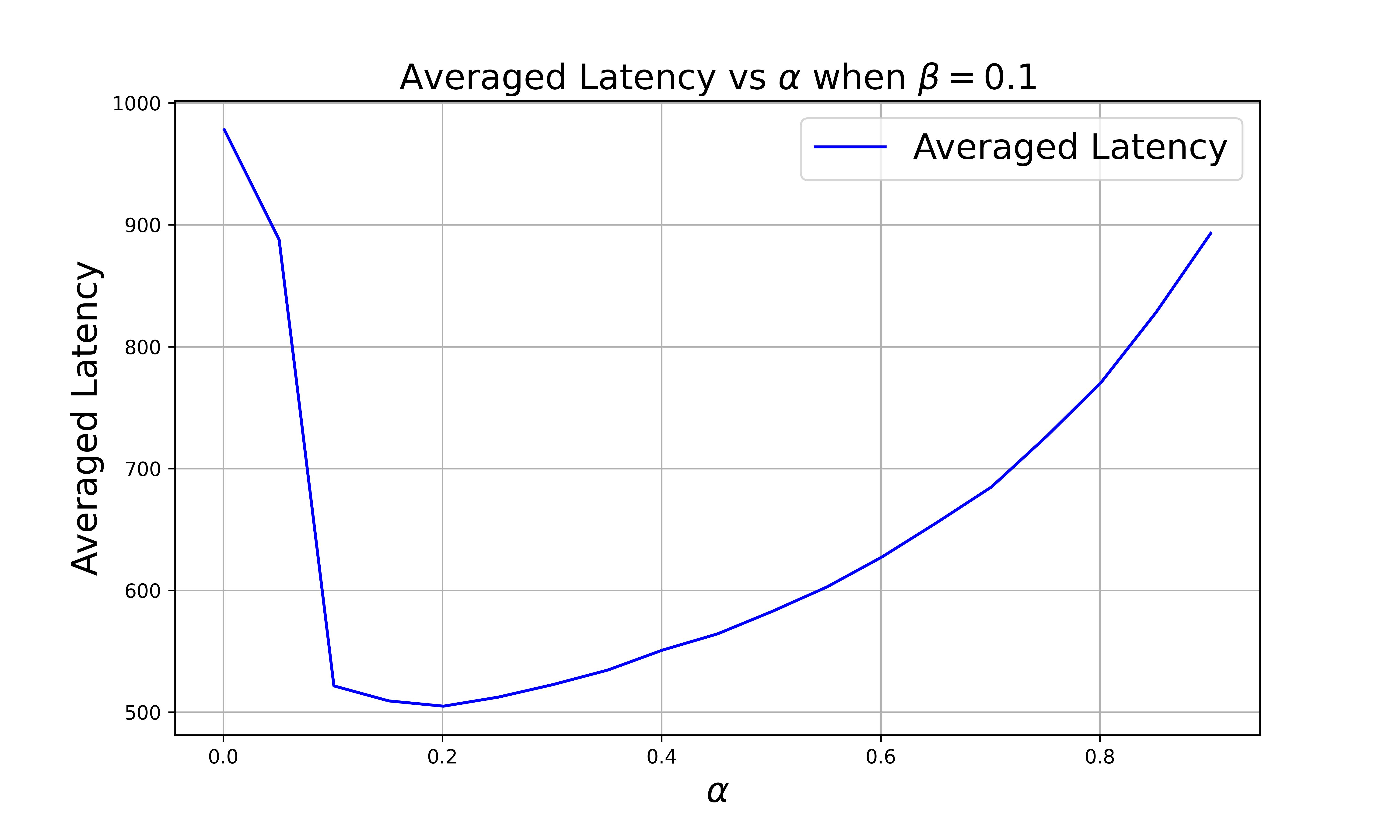}
\includegraphics[width=0.45\textwidth]{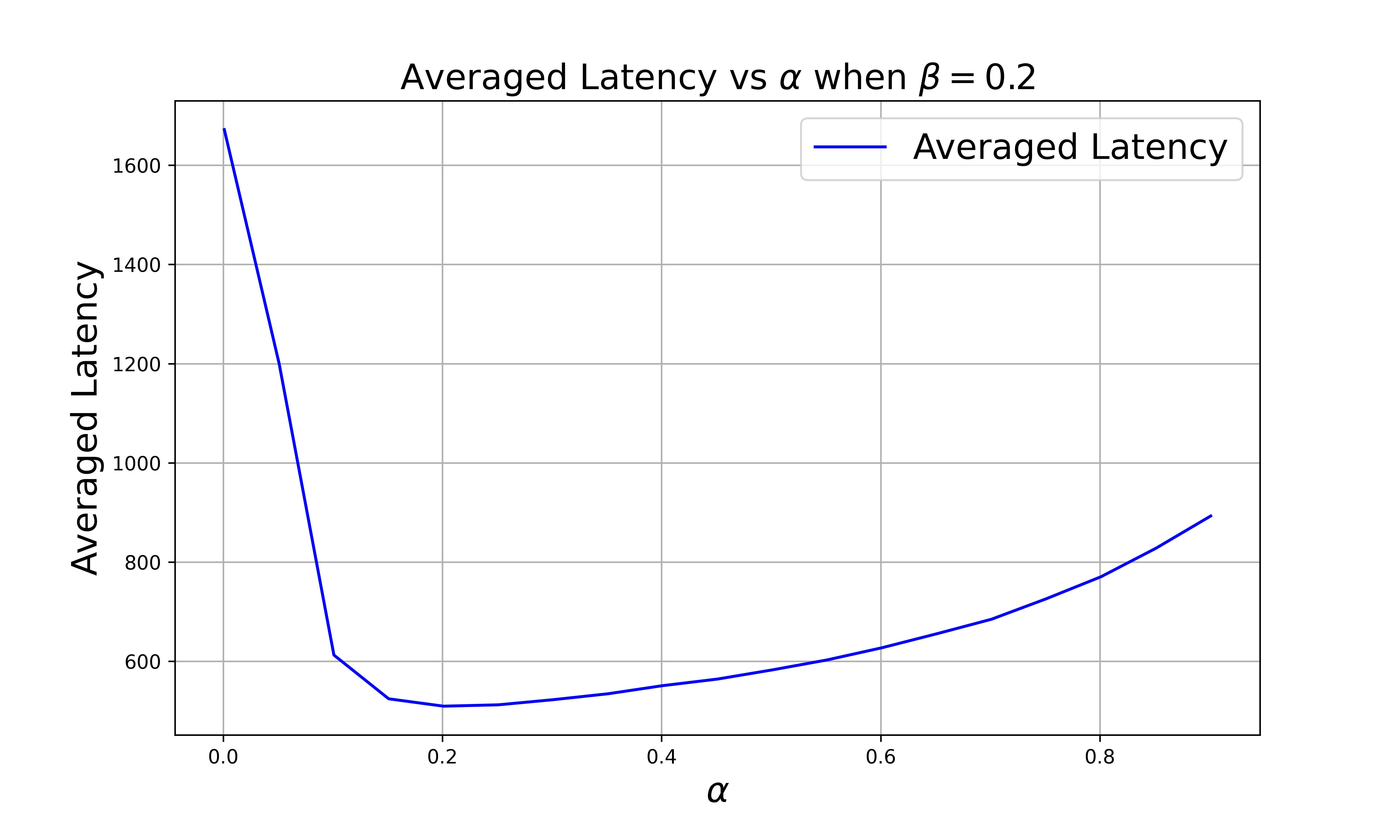}
\caption{Average End-to-End Latency for Different $\alpha$ Values under the High Demand Case} 
\label{fig:alphahigh}
\end{figure}

\begin{figure}[!tb]
\center
\includegraphics[width=0.45\textwidth]{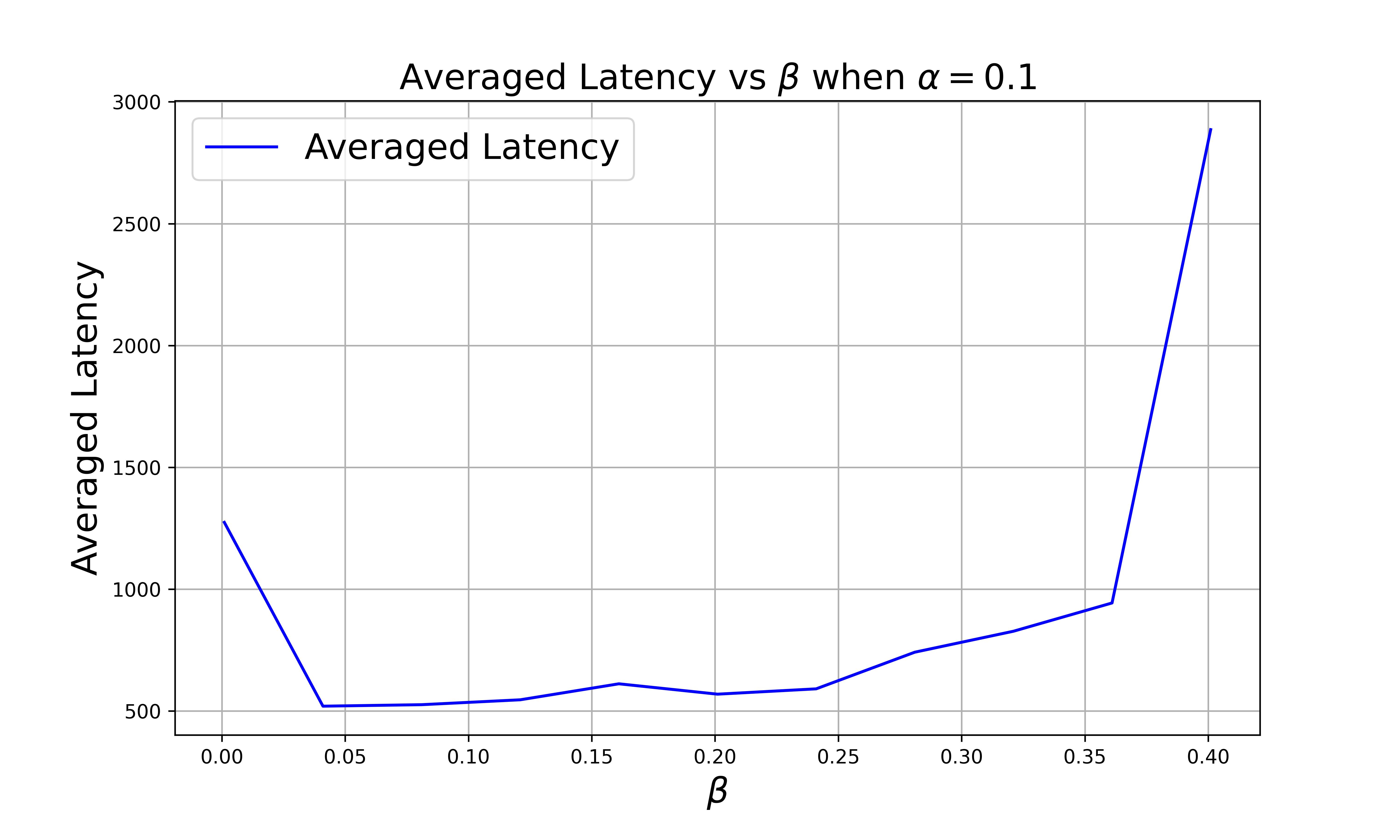}
\includegraphics[width=0.45\textwidth]{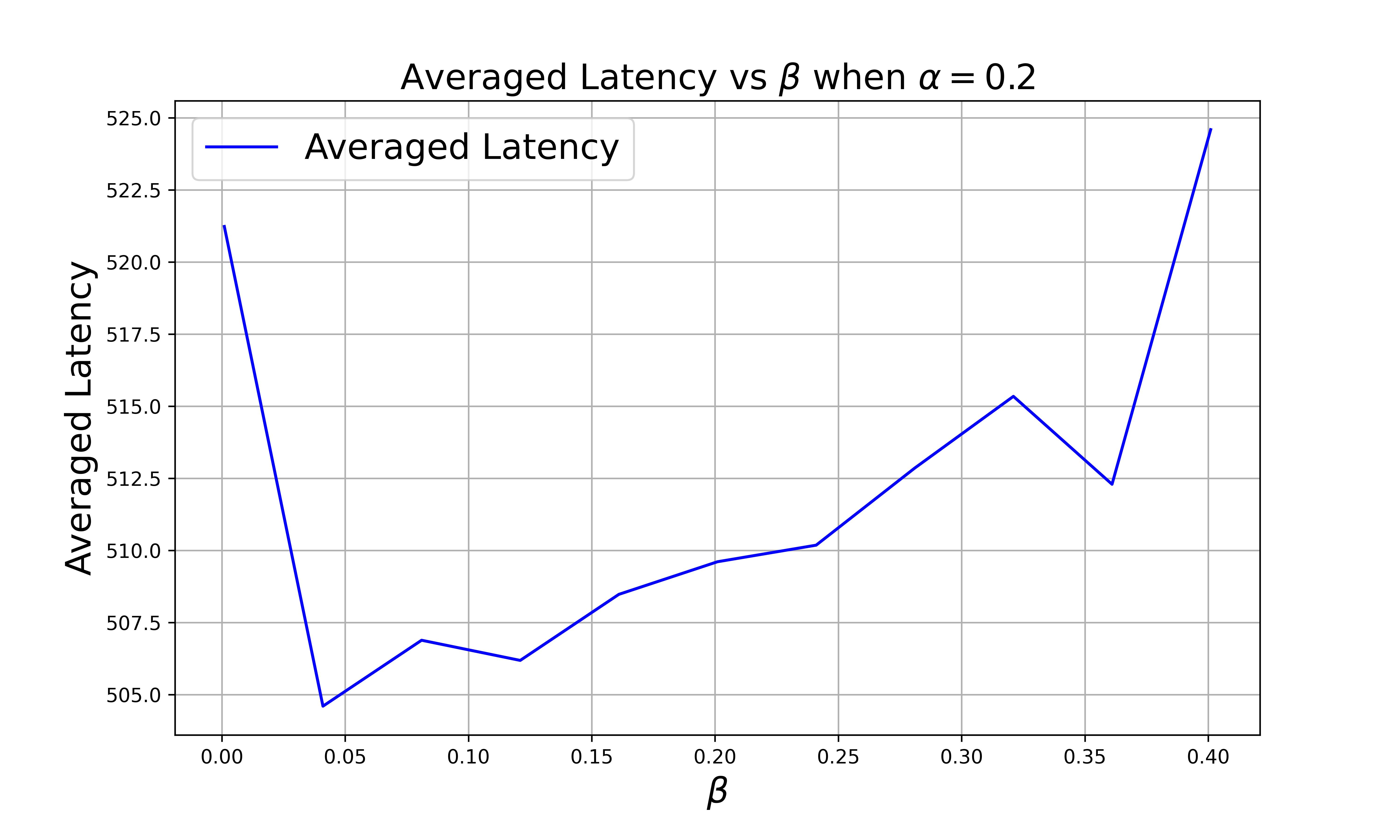}
\caption{Average End-to-End Latency for Different $\beta$ Values under the High Demand Case} 
\label{fig:betahigh}
\end{figure}

\begin{figure}[!tb]
\center
\includegraphics[width=0.8\textwidth]{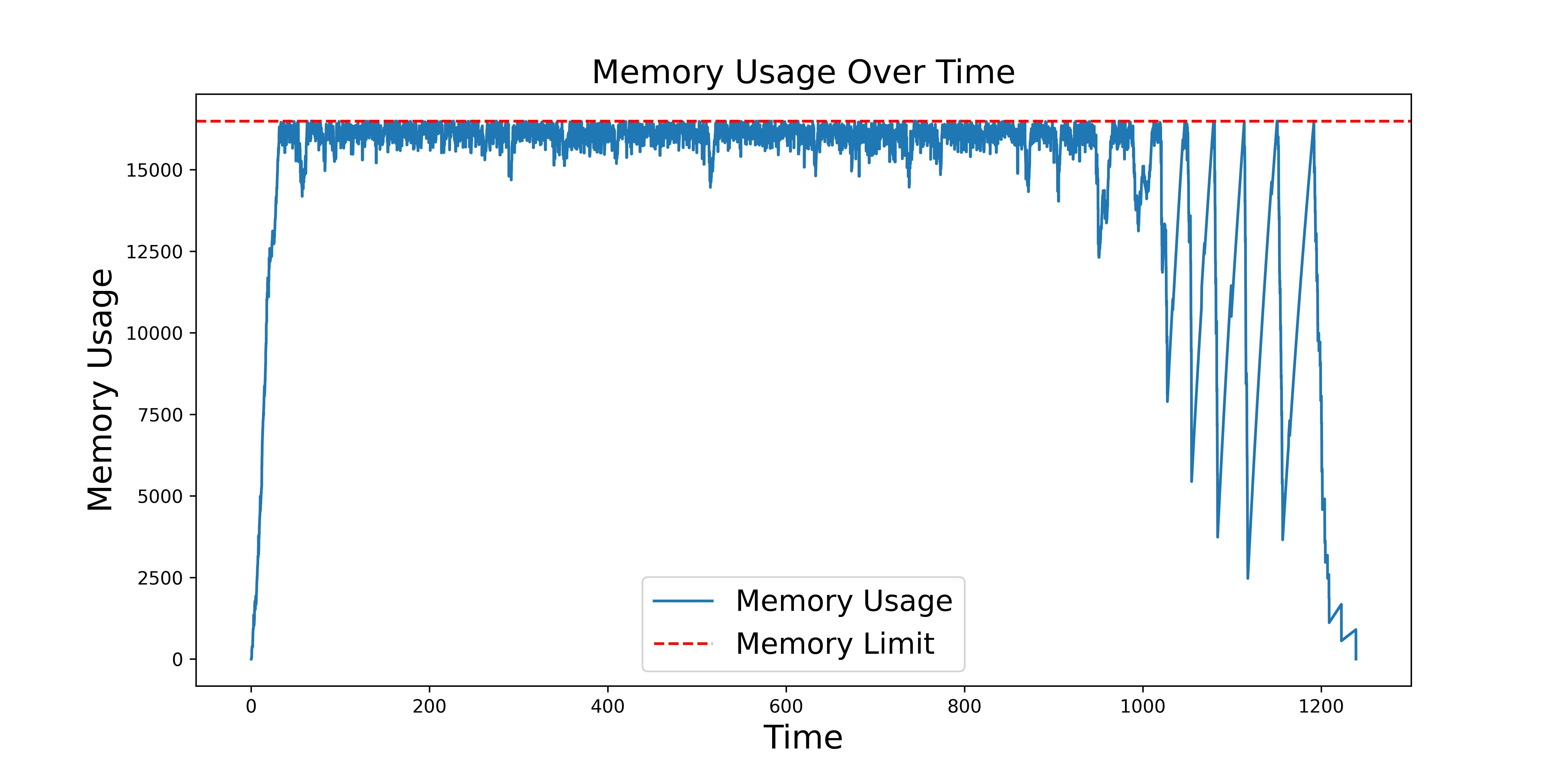}
\caption{Memory Usage over Time for Algorithm~\ref{alg:1} in the Low Demand Case} 
\label{fig:Lmovertime}
\end{figure}

\begin{figure}[!tb]
\center
\includegraphics[width=0.45\textwidth]{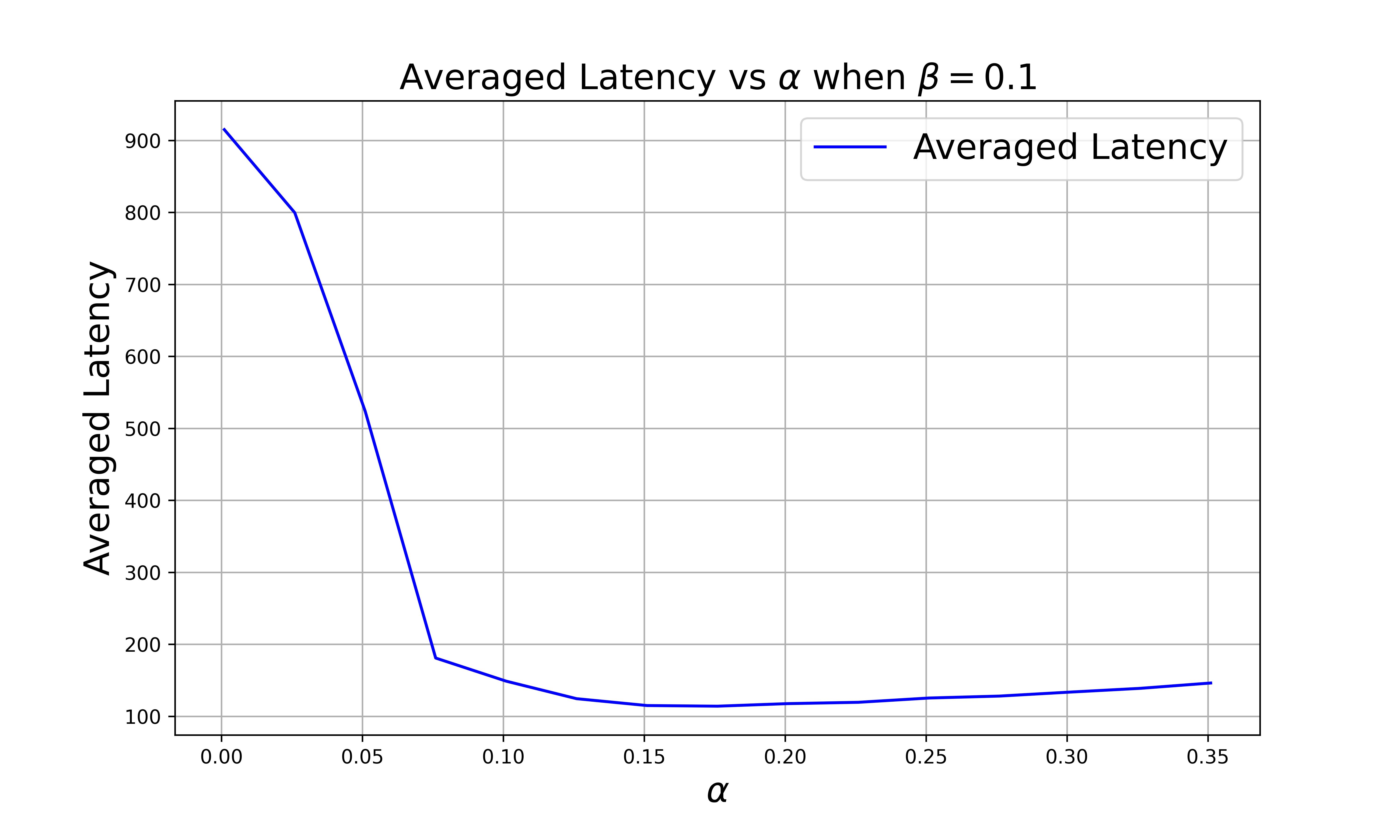}
\includegraphics[width=0.45\textwidth]{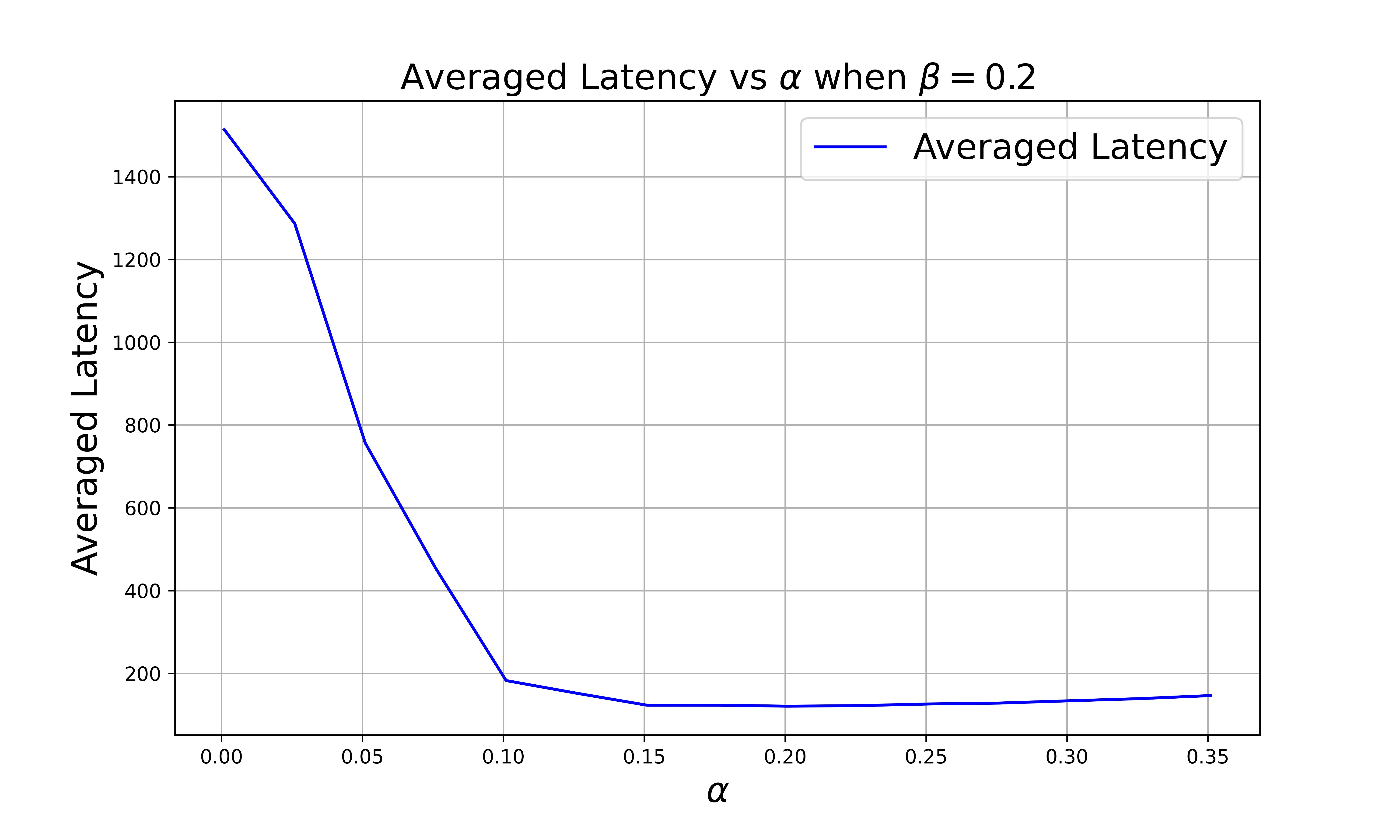}
\caption{Average End-to-End Latency for Different $\alpha$ Values under the Low Demand Case} 
\label{fig:alphalow}
\end{figure}

\begin{figure}[!tb]
\center
\includegraphics[width=0.45\textwidth]{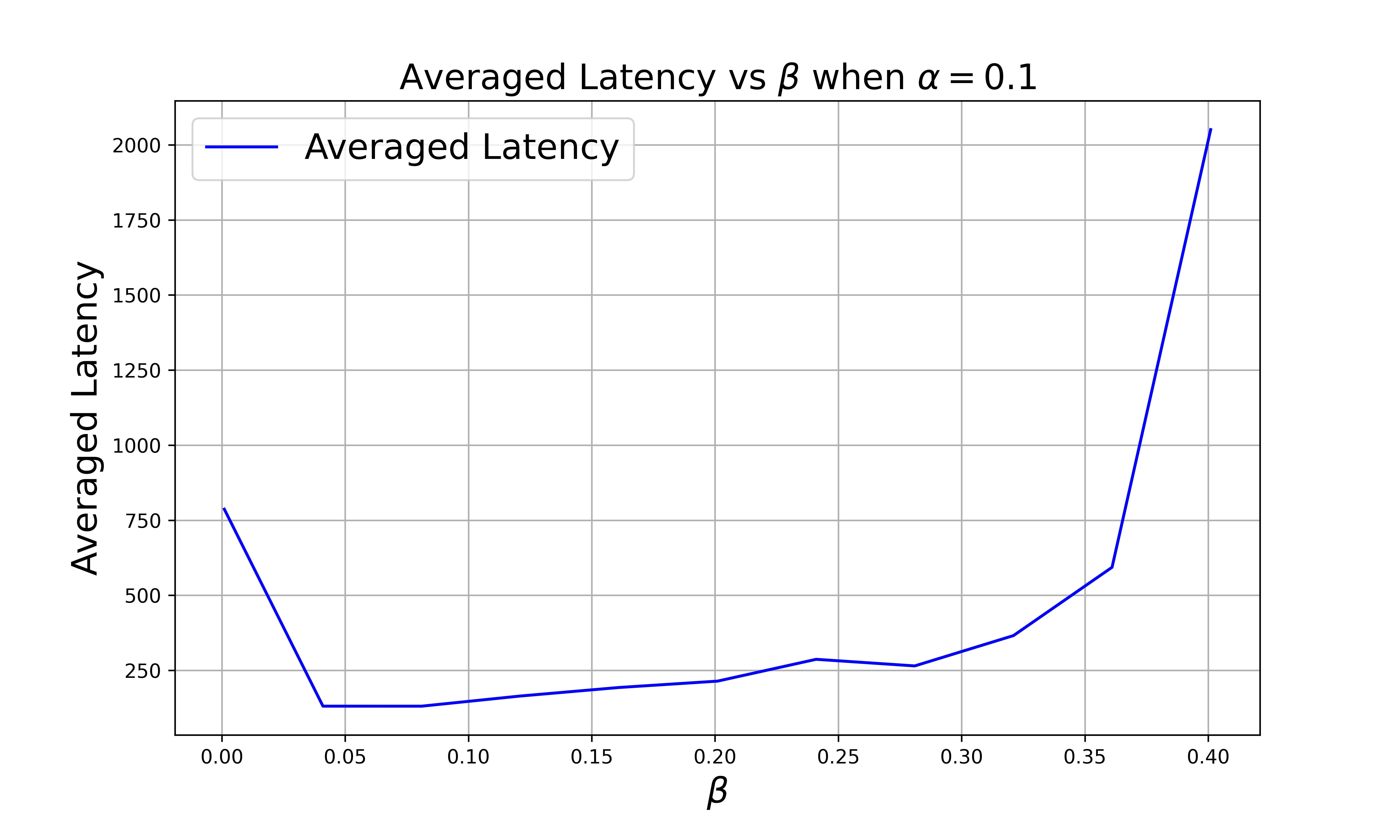}
\includegraphics[width=0.45\textwidth]{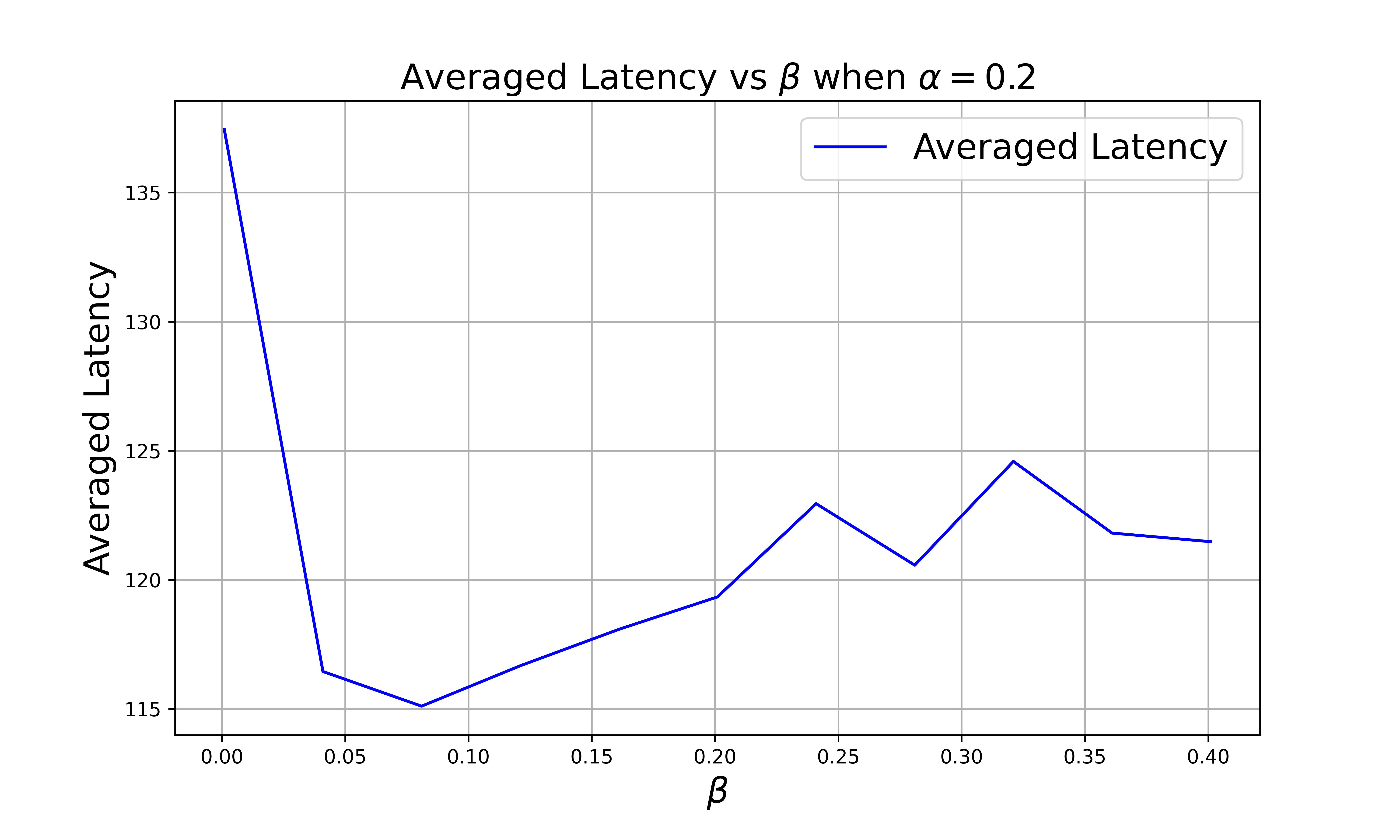}
\caption{Average End-to-End Latency for Different $\beta$ Values under the Low Demand Case} 
\label{fig:betalow}
\end{figure}

We next summarize the figures that support our key findings in the numerical experiments. Figure~\ref{fig:distribution} displays the empirical distributions of input prompt lengths and output response lengths from the dataset used in our study. 
Figures~\ref{fig:movertime} and~\ref{fig:Lmovertime} show that our algorithm consistently utilizes most of the available memory per batch, under both high-demand and low-demand conditions respectively. Figures~\ref{fig:alphahigh} and~\ref{fig:alphalow} report the total latency for various values of the protection level parameter \( \alpha \), and identify recommended ranges for its selection in high- and low-demand scenarios, respectively. Finally, Figures~\ref{fig:betahigh} and~\ref{fig:betalow} evaluate the impact of the KV cache clearing probability \( \beta \) on total latency, highlighting appropriate ranges of \( \beta \) for high- and low-demand settings respectively.

\begin{figure}[!h]
\center
\includegraphics[width=0.8\textwidth]{input_output_distribution.jpg}
\caption{Distribution of the number of words of input prompt and output response respectively} 
\label{fig:distribution}
\end{figure}

In Figure \ref{fig:alphahigh}, we varied $\alpha$ while fixing $\beta=0.1$ and $\beta=0.2$. For both settings, $\alpha \in [0.15,0.25]$ minimizes average latency. When $\alpha < 0.1$, performance degrades significantly as the protected memory is insufficient, necessitating frequent clearing and rescheduling of requests, which leads to redundant computation.

Figure \ref{fig:betahigh} illustrates average latency for various values of $\beta$ with fixed $\alpha$ values of $0.1$ and $0.2$. In both cases, the algorithm performs well with $\beta \in [0.05, 0.25]$. For extremely low $\beta$, the algorithm underperforms as insufficient clearing of requests limits memory availability. This may keep the memory usage be above the limit after clearing the processing requests. To clear enough memory with extremely small $\beta$, this requires a significant time. Conversely, higher $\beta$ values are inefficient due to excessive request clearing, resulting in increased recomputation.

In Figure \ref{fig:alphalow}, we vary $\alpha$ while fixing $\beta$ at $0.1$ and $0.2$. For both settings, values of $\alpha$ in the range $[0.10,0.25]$ minimize average latency, while $\alpha < 0.1$ leads to significant performance degradation.

Figure \ref{fig:betalow} shows the average latency across various $\beta$ values with $\alpha$ fixed at $0.1$ and $0.2$. In both cases, the algorithm achieves stable performance for $\beta$ values between $0.05$ and $0.20$. These trends are consistent with those observed in the high-demand scenario, indicating that similar parameter tuning benefits both models.

Finally, we provide a table with relevant statistics for the case of $1000$ requests with arrival rate $\lambda = 50$ for $50$ independent runs.

\begin{table}[ht]
\centering
\caption{Relevant statistics among $50$ independent experiments of different algorithms where there are $1000$ requests, and the arrival rate $\lambda=50$}
\begin{tabular}{lcccc}
\toprule
Algorithm & Average & Std. Dev. & Max & Min \\
\midrule
\mcsdf & 32.112 & 0.354 & 33.097 & 31.505 \\
\mcb & 46.472 & 0.310 & 47.135 & 45.838 \\
Benchmark $\alpha=0.3$ & 51.933 & 0.324 & 52.532 & 51.204 \\
Benchmark $\alpha=0.25$ & 51.046 & 0.351 & 51.757 & 50.279 \\
Benchmark $\alpha=0.2$, $\beta=0.2$ & 50.401 & 0.343 & 51.035 & 49.700 \\
Benchmark $\alpha=0.2$, $\beta=0.1$ & 50.395 & 0.360 & 51.083 & 49.586 \\
Benchmark $\alpha=0.1$, $\beta=0.2$ & 53.393 & 1.457 & 56.357 & 49.488 \\
Benchmark $\alpha=0.1$, $\beta=0.1$ & 50.862 & 0.946 & 53.978 & 49.086 \\
\bottomrule
\end{tabular}
\label{tab:performance}
\end{table}

\newpage